\documentclass[11pt]{article}

\usepackage[T1]{fontenc}
\usepackage[utf8]{inputenc}
\usepackage[a4paper,margin=1in]{geometry}
\usepackage{newtxtext}
\usepackage{newtxmath}
\usepackage{microtype}
\usepackage{authblk}
\usepackage{bm}
\usepackage{amsmath}
\usepackage{mathtools}
\usepackage{graphicx}
\usepackage{booktabs}
\usepackage{algorithm}
\usepackage{algpseudocode}
\usepackage{tikz}
\usepackage{xcolor}
\usepackage[most]{tcolorbox}
\usepackage{wrapfig}
\usepackage[sort&compress,numbers]{natbib}
\usepackage{url}
\usepackage[colorlinks=true,linkcolor=blue!45!black,citecolor=blue!45!black,urlcolor=blue!45!black]{hyperref}

\usetikzlibrary{spy}

\definecolor{accentcolor}{HTML}{0ab246}

\tcbset{colback=gray!5!white,colframe=gray!75!black,boxrule=0.5pt,arc=3pt}

\bibpunct[, ]{[}{]}{,}{n}{}{,}

\title{\bfseries Foundations of Reinforcement Learning and Control:\\ Connections and New Perspectives}

\author[1,*]{Claire Vernade}
\author[2,*]{Onno Eberhard}
\author[3]{Martha White}
\author[4]{Florian D\"orfler}
\author[3]{Csaba Szepesv\'ari}
\author[5]{\mbox{Miroslav Krstic}}
\author[2]{Michael Muehlebach}

\affil[1]{University of Technology Nuremberg, \texttt{claire.vernade@utn.de}}
\affil[2]{Max Planck Institute for Intelligent Systems, T\"ubingen, Germany, \texttt{oeberhard@tue.mpg.de}, \texttt{michaelm@tue.mpg.de}}
\affil[3]{University of Alberta and Alberta Machine Intelligence Institute (Amii), \texttt{whitem@ualberta.ca}, \texttt{szepesva@ualberta.ca}}
\affil[4]{ETH Z\"urich}
\affil[5]{University of California San Diego}
\affil[*]{Equal contribution.}

\date{}

\begin{document}

\maketitle

\begin{abstract}
Reinforcement learning and control theory are two adjacent scientific fields that focus on optimizing the controller of unknown dynamical systems using feedback. While both fields have common roots in dynamic programming, they have evolved with distinct methodologies, goals, and cultures. Despite decades of mutual influence, a significant gap persists between the two communities. This tutorial introduces adaptive control, actor-critic reinforcement algorithms, and a new way to combine these two paradigms for data-driven decision making on a classical locomotion control problem. Our aim is to provide a foundation for understanding the core differences between the two approaches and insights to help experts in each field better understand and engage with the tools and approaches of the other.
\end{abstract}

\medskip
\noindent\textbf{Keywords:} reinforcement learning; control theory; adaptive control; dynamic programming; actor-critic methods

\bigskip

\section{Introduction}

Dynamical systems are ubiquitous in physical, biological and digital environments, and controlling them towards given targets has been an ongoing challenge for more than a century. Control theory and reinforcement learning (RL) are now at the forefront of this challenge, and are active fields of research tackling data-driven decision making. Despite fundamentally different histories, the similarity of the challenges has recently pushed the convergence of the two communities, emphasizing adaptation, universality and broad applicability of the produced methods.

Adaptive control emerged in the 1960s as a powerful response to a central engineering challenge: how to maintain stability and performance when the dynamics of a system are uncertain or change over time \cite{astromHistory,annaswamy2021historical}. Motivated by practical needs in aerospace, manufacturing, and process control, the field quickly gained prominence. Early experiments, such as gain-scheduled controllers for high-performance aircraft \cite{aircraftAdaptive1961} and self-tuning regulators for industrial plants \cite{caldwell1950control,gabor1961universal}, demonstrated that it was possible to design feedback systems that could adjust in real time to uncertain or evolving dynamics. These successes generated much optimism: adaptive systems appeared to behave ``intelligently'' by tuning themselves in real time, offering promising performance in uncertain and time-varying environments. The theoretical foundations developed in this period, including model-reference adaptive control (MRAC) \citep{narendra1989stable,ioannou2012robust} and adaptive pole placement, laid the groundwork for decades of research.

With the start of the 1990s, the attention in adaptive control turned from linear systems to nonlinear systems. Fuelled by the structural advances in geometric nonlinear control, the highly cited adaptive backstepping \citep{krstic1995nonlinear} provided parameter-adaptive Lyapunov-based designs for feedback-linearizable systems, as well as systems beyond this class, with unlimited amounts of parametric uncertainty. The application range of adaptive nonlinear control is broad: robotics, vehicles, electric machines, power electronics, propulsion and flows, and numerous other applications.

In parallel, in the 1980s, neural networks \cite{rosenblatt1958perceptron} emerged in artificial intelligence as complex families of parametrized functions that could be tuned using data, provided that the optimization could be performed in an efficient way. This learning mechanism can also be viewed as a dynamical system, where current parameter values are updated sequentially using available data.
The similarity with brain development prompted cognitive scientists to search for optimization algorithms that would resemble the processes occurring in the human brain.
One of the key mechanisms at stake in this natural optimization process is the reinforcement of connections through the backward propagation of dopamine signals. This idea inspired the \emph{temporal difference} algorithm \citep{sutton1988learning}, originally aimed at optimizing neural networks but soon connected to adaptive control \citep{sutton2002reinforcement} and approximate dynamic programming \citep{bertsekas1996neuro}, giving rise to the research field now known as \emph{reinforcement learning}.

In recent years, both RL and control have evolved due to advances in data availability and computational infrastructure. This new technological environment has led both fields to converge. A growing community of control researchers now approach control problems from a data-centric, algorithmic, and computational perspective \citep{dorfler2023data}.
In parallel, RL researchers are now increasingly working on physical systems, aiming at exploiting domain knowledge and providing stronger guarantees for downstream users \citep{dulac2021challenges,xie2025safe}.

These developments have enabled the study and deployment of control strategies in complex, high-dimensional environments, where challenges go well beyond classical notions of regulation or stabilization \cite{tang2025deep,elmkaiel2025embodied,liu2022stability,ma2023reinforcement,he2025decision,piazza2019century}. Modern applications demand controllers capable of handling non-equilibrium motion, non-smooth dynamics, and changing goals in real time. Such demands are increasingly common in domains like robotic locomotion, manipulation, and interactive decision making---areas that continue to push the limits of both learning and control.

Recent progress points to a renewed opportunity to revisit and significantly expand the core ambitions of adaptation and learning set out by adaptive control. With the convergence of data-driven algorithms, robust control theory, and scalable computation, the time is ripe to bring together the RL and control communities. This tutorial contributes to that effort by clarifying how each field conceptualizes the problem of learning to control unknown systems, identifying their complementary strengths, and outlining a path toward more unified and practically effective approaches.

\subsubsection*{Structure of the tutorial. } We aim at providing a concise but technical introduction to both research fields to allow newcomers as well as researchers from either side to join efforts.

The tutorial starts with a general introduction (Section~\ref{sec:problem_definition}) to the learning problem, introducing a unified notation and emphasizing the common principles that underlie both RL and control, namely dynamical systems, Lyapunov functions and dynamic programming. We introduce a locomotion task as a running example to illustrate the methods throughout the tutorial. Additionally, we discuss in more detail the methodological differences between the fields and how they form two avenues toward the common problem of \emph{data-driven decision making}.

The core technical section of this tutorial then explains in detail two main approaches that we chose as representative methodologies from the two fields. We first discuss \emph{model-reference adaptive control} (MRAC, Sec.~\ref{subsec:adaptive_control}, with extensions in Sec.~\ref{sec:beyond_mrac}), where data-driven adaptivity is added on top of a base model of the system. In contrast, we introduce \emph{actor-critic} algorithms (Sec.~\ref{sec_rl_alg}) as major representatives of model-free reinforcement learning. We then present an original and successful combination of these two approaches, using an actor-critic algorithm for high-level learning and planning and MRAC for low-level adaptation (Sec.~\ref{subsec:combining_RL_and_control}). We also discuss the distinction between model-based and model-free algorithms in RL and blur the standard boundaries to further highlight connections with control.

\section{Problem definition}
\label{sec:problem_definition}
At their heart, both RL and control theory are concerned with the general problem of sequentially selecting actions to make a dynamical system achieve a desired behavior.
In this section, we formalize the problem setting we consider in this article, and clarify the different terminology used by control theorists and reinforcement learners.
We begin by introducing the \emph{Half-Cheetah} system, a simulated locomotion task, which will be a running example for the rest of the tutorial.
We then present the foundations of RL and adaptive control, starting with a discussion of the common foundation on which both fields are based: dynamical systems and optimal control.

\subsection{Running example: the Half-Cheetah system}
\label{subsec:prob_formulation}
\begin{wrapfigure}[9]{r}{5.4cm}
\vspace{0.25\baselineskip}
\centering
\pgfdeclarelayer{fg}
\pgfsetlayers{main,fg}
\begin{tikzpicture}[x=1cm, y=1cm,
    spy using outlines={circle, accentcolor, magnification=4, size=2cm, connect spies}
]
    \node[draw, inner sep=0] (image) at (0,0) {
            \includegraphics[width=3cm]{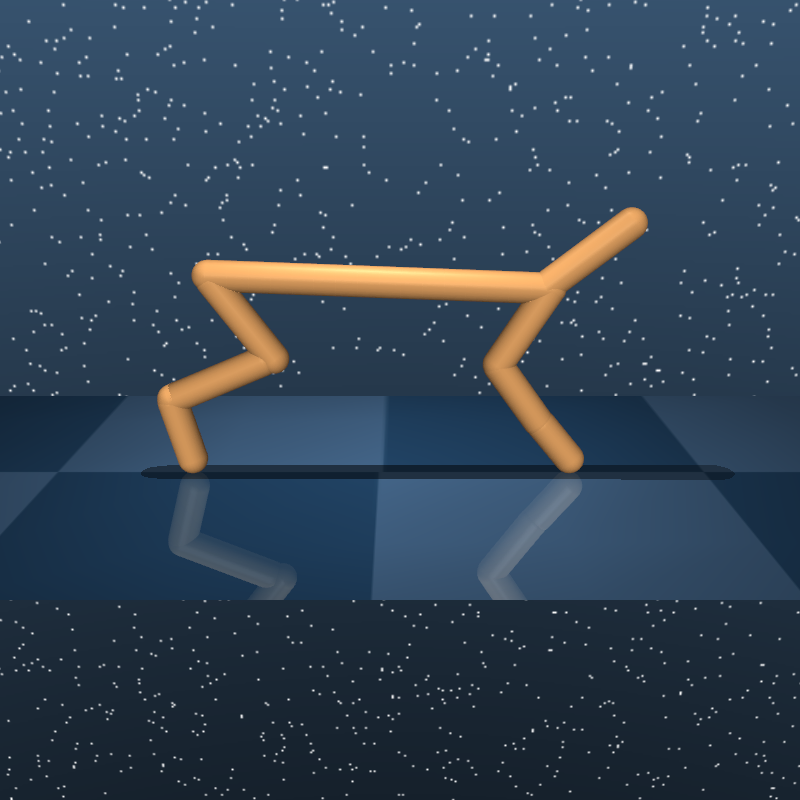}
    };
    \spy on (0.45, 0.1) in node at (2, 0.5);
    \begin{pgfonlayer}{fg}
    \begin{scope}[xshift=1.96cm, yshift=0.61cm]

        \draw[white, thick] (0, 0) ++(60:.2) arc (60:-55:.2);
        \node[white] at (0.4, 0.05) {$\theta$};
    \end{scope}
    \end{pgfonlayer}
\end{tikzpicture}
\vspace{0.25\baselineskip}

\end{wrapfigure}
The Half-Cheetah, shown on the right, is a simplified robotic system simulated in MuJoCo \citep{todorov2012mujoco}.\footnote{We use the DeepMind Control Suite implementation \citep{tassa2018deepmind,zakka2025mujocoplayground}. The image of the Cheetah has been reproduced with permission from \citep{tassa2018deepmind}.}
This system is the basis of a common benchmark task for RL algorithms, with the goal of making the robot `run' as far as possible within a given time limit.

In Section~\ref{sec_rl_alg}, we illustrate how a modern deep RL algorithm, \emph{Soft Actor-Critic} (SAC), can learn a locomotion policy directly from interaction with the environment. In this setting, the objective is to maximize performance---for example, making the robot run as fast as possible---without imposing a reference trajectory that the robot must follow.

By contrast, robotic systems such as the one considered here have long been studied in control theory, where the objective is often formulated in terms of stability or trajectory tracking.
For example, the Half-Cheetah system can be decomposed into six individual joints whose angles $\theta$ are used to describe the system.
Though the governing equations are complex and highly nonlinear, the behavior of these joints can be approximately described by independent second-order linear dynamical systems:
\begin{equation}
\label{eq:sos}
    \ddot\theta + 2\zeta\omega\dot\theta + \omega^2\theta = \frac{\tau}{I} + \eta,
\end{equation}
where $\zeta$ is a damping (friction) coefficient, $\omega$ is the natural angular frequency of the joint, $\frac{\tau}{I}$ is the applied torque divided by the rotational inertia, and $\eta$ is a ``disturbance'' term that captures all nonlinear effects.
This is a useful model if $\eta$ is small, and in Section~\ref{subsec:adaptive_control}, we show how MRAC enables reference tracking in such a second-order system, even if model parameters are unknown.

RL and control theory traditionally see control problems through different lenses, and in this article we argue that these different points of view are often complementary.
As an illustration, we develop an original method for combining high-level RL with low-level MRAC.
In Section~\ref{subsec:combining_RL_and_control}, we test this approach in the Half-Cheetah environment, and show that it enables both high performance (thanks to the actor-critic algorithm) and continual adaptation to changing low-level dynamics (thanks to the adaptive controller).

Beyond the Half-Cheetah benchmark, similar architectures are natural whenever a complex system admits a useful separation between high-level decision making and low-level stabilization. Examples include legged robots operating on changing terrain, aerial or underwater vehicles with uncertain payloads or fluid interactions, robotic manipulators whose dynamics change with the grasped object, assistive devices such as prostheses or exoskeletons that must adapt to the user, and chemical processing plants that operate under varying loads and feedstock compositions. In such settings, RL can be used to learn task-level behavior from experience, while MRAC can provide continual adaptation of the low-level tracking controllers.

\subsection{Preliminaries: dynamical systems}
\label{subsec:dynamics}

RL and control theory have developed independently, and this is reflected in the notation and assumptions commonly used, even in their treatment of the fundamental object underlying both fields: the dynamical system. We provide a unified language and notation (summarized in Appendix~\ref{app:notation}) and highlight that RL and control are fundamentally concerned with the same problem: designing a controller, or \emph{policy}, that drives a dynamical system towards a goal.

\noindent
A dynamical system is a model mainly composed of three elements:
\begin{itemize}
    \item A state space $\mathcal{S}$: this is a set of variables that describes the world the policy, or controller, is interacting with;
    \item An action space $\mathcal{A}$: these are the controls that are available to the policy to impact the environment, i.e., modify the state;
    \item A dynamics function that dictates how the state changes as a function of the current state and the action chosen by the learner. The dynamics may or may not depend on time, and the two fields have adopted different notations (see box below).
\end{itemize}

In RL, the standard modeling formalism is a \emph{controlled Markov chain}, or Markov decision process, induced by a kernel that defines state transitions conditionally on actions.
In control, it is more common to model the dynamics of a physical system with a deterministic function and an additive noise term representing external perturbations. This can be summarized as follows:

\vspace{0.5cm}
\noindent

\begin{tcolorbox}

\begin{minipage}[t]{0.48\textwidth}
\textbf{RL view.}

RL adopts a stochastic formulation based on controlled Markov chains.
At round $t$, given a state $s_t \in \mathcal{S}$ and an action $a_t \in \mathcal{A}$, the next state evolves as
\begin{equation*}
  s_{t+1} \sim p(\,\cdot \mid s_t,a_t),
\end{equation*}

where $p(\cdot \mid s,a)$ is the distribution of the next state.
\end{minipage}
\hfill
\begin{minipage}[t]{0.48\textwidth}
\textbf{Control view.}

Control is typically formulated for physical systems with deterministic dynamics subject to disturbances.
The state evolution is written
\[
  s_{t+1} = f(s_t,a_t) + \eta_t,
\]
where $f$ is a (possibly known) map and $\eta_t$ represents external perturbations.
\end{minipage}
\end{tcolorbox}

While the underlying problem is shared, the two fields differ in their formulations and solution methods. A central aim of this tutorial is to bridge these perspectives. We therefore begin by introducing the key concepts, objectives, and tools used in each community.

\subsection{Stability and robustness}
\label{subsec:stability_lyapunov}
A central objective in control is to design policies that ensure that the dynamical system behaves in a predictable and safe manner over time. A natural first approach is to input a fixed sequence of actions into the system independently of observed states; this is called \emph{open-loop control}.
However, to achieve stability, a feedback policy is often needed: given a target state $s^\star \in \mathcal{S}$, one seeks a policy $\pi$ that takes an observation of the state as input (called \emph{closed-loop control}) such that the resulting trajectory $(s_t)_{t \geq 0}$ remains close to $s^\star$, or converges to it. In the deterministic setting, this amounts to requiring that $s^\star$ is an equilibrium of the closed-loop system,
\[
f(s^\star, \pi(s^\star)) = s^\star,
\]
and that trajectories initialized near $s^\star$ remain close (stability) or converge to it (asymptotic stability). In the presence of stochastic perturbations, these requirements are relaxed to hold in expectation or with high probability, leading to notions of robustness with respect to disturbances.

A standard tool to analyze such stability properties is the notion of a \emph{Lyapunov function}. A function $U : \mathcal{S} \to \mathbb{R}_+$ is called a Lyapunov function for a policy $\pi$ around $s^\star$ if
\begin{align*}
    &U(s^\star) = 0, \quad U(s) > 0 \quad \text{for all } s \neq s^\star, \\
    &U\big(f(s,\pi(s))\big) - U(s) \leq 0 \quad \text{for all } s \in \mathcal{S}.
\end{align*}
The function $U$ can be interpreted as a measure of distance or energy with respect to the target state, and the second condition enforces that this quantity decreases along trajectories. Stronger conditions, such as a strict decrease away from $s^\star$, imply asymptotic convergence to the target state.

Lyapunov functions provide a powerful and general framework. However, they are not given a priori. Constructing a suitable function $U$ is hard work, demanding skill and experience, accumulated over a large body of literature \citep{Khalil2002}. Since the work of \citet{krstic1995nonlinear}, much of the adaptive control literature---comprising thousands of papers over the past three decades---has focused not only on the design of adaptive controllers, but more fundamentally on the construction of Lyapunov functions guaranteeing stability.

\subsection{Optimal control \& decision-making}
\label{subsec:opt_control}

\paragraph{The linear-quadratic control approach: a first value function. }

While there is no general recipe for arbitrary systems, the case of linear dynamical systems is well understood and provides a pedagogical example of a situation where Lyapunov functions can be obtained through an optimal control objective. Consider a linear dynamical system
\[
s_{t+1} = A s_t + B a_t,
\]
where the target is $s^\star=0$, and $A$ and $B$ are given.\footnote{In adaptive control, a prescribed pair of matrices can instead define the reference model that guides adaptation based on feedback (see Section~\ref{subsec:adaptive_control}).} We aim to construct a linear policy $a_t = K s_t$, which simplifies the closed-loop dynamics into $s_{t+1} = (A + BK)s_t$. A natural candidate Lyapunov function $U$ is quadratic: for some choice of $P \succ 0$,
\[
U(s) = s^\top P s.
\]
Certifying stability for our controller $K$ then boils down to ensuring that

\[
(A+BK)^\top P (A+BK) - P \preceq 0.
\]

This reduces the problem of verifying stability to checking a matrix inequality, and forms the basis of much of linear control theory \cite{Boyd1994}.

However, rather than constructing such a Lyapunov function directly, an alternative perspective is to define an optimization objective and derive both the policy and the Lyapunov function from it. This is the viewpoint adopted in the linear quadratic regulator (LQR) formulation, where one considers the cost
\[
\sum_{t=0}^{\infty} \big( s_t^\top Q s_t + a_t^\top R a_t \big),
\]
for some known matrices $Q \succeq 0$ and $R \succ 0$ chosen by the designer. These matrices typically encode a trade-off between penalizing deviations of the state from the target and limiting the magnitude of the control actions. The goal is then to find a policy minimizing this cumulative cost.

More generally, given a sequence of actions $(a_t)_{t\geq0}$ output by a controller $\pi$, the \emph{cost-to-go function} (or negative \emph{value function}) associated with this objective is defined as the expected cumulative cost when starting from a state $s$:
\[
J^\pi(s) = \mathbb{E}\left[ \sum_{t=0}^{\infty} \big( s_t^\top Q s_t + a_t^\top R a_t \big) \,\middle|\, s_0 = s \right],
\]
where the expectation is taken over the trajectories induced by the dynamics and the policy. The value function therefore quantifies the long-term performance of a policy from any initial state.

In this setting, it can be shown, e.g., by dynamic programming (explained further below), that the optimal policy is linear, and the associated optimal cost-to-go takes the quadratic form
\[
J^\star(s) = s^\top P^\star s,
\]
where $P^\star$ is the solution of the algebraic Riccati equation \citep{AndersonMoore1971,Bertsekas2017}.

Cost-to-go functions and Lyapunov functions originate from different perspectives. The former quantify the long-term performance with respect to a given objective while the latter certify stability of an equilibrium. However, for LQR (and beyond, as we shall see soon), a remarkable alignment occurs: the optimal cost-to-go function $J^\star$ is also a Lyapunov function for the closed-loop system.
Indeed, along trajectories of the optimal policy,
\[
J^\star(s_{t+1}) - J^\star(s_t) = - \big( s_t^\top Q s_t + a_t^\top R a_t \big) \leq 0,
\]
so that the decrease of the cost-to-go exactly matches the instantaneous cost. As a consequence, optimality and stability are achieved simultaneously.

This example illustrates a broader perspective: Lyapunov functions provide a direct way to enforce stability through local descent conditions.
This viewpoint naturally connects to dynamic programming and RL, which we discuss next.

\subsubsection*{Markov Decision Processes and Dynamic Programming. }

To simplify the exposition, we focus here on discounted infinite-horizon problems. Given a discount factor $\gamma \in [0,1)$, the goal is to find a feedback controller, or \emph{policy}, $\pi:\mathcal{S}\to \Delta(\mathcal{A})$ that minimizes the expected cumulative cost-to-go function, where $\Delta(\mathcal{A})$ denotes the set of probability distributions over actions:
\begin{equation}
\label{eq:main-objective}
    J(\pi)
    \;=\;
    \mathbb{E}
    \left[
      \sum_{t=0}^{\infty} \gamma^t \, c(s_t, a_t)
    \right]
    \;=\;
    -\mathbb{E}
    \left[
      \sum_{t=0}^{\infty} \gamma^t \, r(s_t, a_t)
    \right] = -V(\pi),
\end{equation}
where the expectation is taken over the system dynamics, the possible stochasticity of the policy, and the initial condition $s_0 \sim p_0$.
Here, we have introduced the RL terminology of \emph{rewards} $r$ (negative costs), and \emph{value functions} $V$ (reward-to-go).
\begin{tcolorbox}
    \begin{minipage}[t]{0.48\textwidth}
    \textbf{RL view. }
    The objective in RL is to maximize cumulative rewards along a trajectory. When rewards are defined as negative costs, $r(s,a)=-c(s,a)$, the reward-to-go is called the \emph{value function} and must be maximized.
    \end{minipage}
\hfill
    \begin{minipage}[t]{0.48\textwidth}
    \textbf{Control view. }
    In control, the objective is most often to reach a zero-cost state, which corresponds to the target. Thus, costs are often assumed to be positive, and $J$ must be minimized by the policy.
    \end{minipage}

\vspace{0.3cm}
    \textbf{In this tutorial, we adopt the RL terminology and focus on maximizing value functions. }

\end{tcolorbox}

How can we efficiently compute an optimal policy, if the action taken in a state $s$ affects not only the immediate reward, but also all future states visited by the system?
The answer to this computational question is given by \emph{dynamic programming} \citep{bellman1957dynamic}, which relies on the recursive structure of the value functions associated with our Markov decision process. Given a policy $\pi$ and a state $s \in \mathcal{S}$, the value function of $\pi$ is defined as the expected downstream rewards when starting in state $s$ and following $\pi$:
\begin{equation}
    \label{eq:value-function}
    V^{\pi}(s)
    \;=\;
    \mathbb{E}\left[
        \sum_{t=0}^{\infty} \gamma^t r(s_t,a_t)
        \,\middle|\,
        s_0=s,\; a_t \sim \pi(\cdot|s_t)
    \right].
\end{equation}
In particular, we have
    $J(\pi) = -\mathbb{E}_{s\sim p_0} \left[ V^\pi(s) \right]$.
Thus, optimizing $J$ amounts to finding a policy whose value function is maximal on the relevant initial states.

By exploiting the Markov property of the dynamics, one obtains the \emph{Bellman equation} for a fixed policy, which connects the value in a given state $s$ with that of the next reachable ones:
\begin{equation}
\label{eq:bellman-policy}
    V^\pi(s)
    =
    \int_{\mathcal{A}} \pi(da|s)
    \left[
        r(s,a)
        + \gamma \int_{\mathcal{S}} V^\pi(s')\, p(ds'|s,a)
    \right].
\end{equation}
Here, $p$ denotes the transition kernel of the controlled Markov process. In the deterministic case where $\pi$ maps each state to a single action, this simplifies to
$
    V^\pi(s)
    =
    r(s,\pi(s))
    + \gamma \int_{\mathcal{S}} V^\pi(s')\, p(ds'|s,\pi(s))
$.
To emphasize the impact of a specific action $a$ in a given state $s$, it is also useful to define the state-action value function, or \emph{$Q$-function}, associated with $\pi$:
\begin{equation}
    \label{eq:Q-function}
    Q^\pi(s,a)
    =
    r(s,a)
    + \gamma \int_{\mathcal{S}} V^\pi(s')\, p(ds'|s,a).
\end{equation}
The value function can then be recovered by averaging the $Q$-function under the policy: $
    V^\pi(s) = \int_{\mathcal{A}} Q^\pi(s,a)\, \pi(da|s)$.

Dynamic programming exploits these recursive equations to improve policies iteratively. Starting from an initial policy $\pi_0$, one alternates between two steps:

\medskip
\noindent
\textbf{Policy evaluation.} Given a policy $\pi_k$, compute its $Q$-function $Q^{\pi_k}$ by solving the Bellman equation
\[
    Q^{\pi_k}(s,a)
    =
        r(s,a)
        + \gamma \int_{\mathcal{S}}\int_{\mathcal A} Q^{\pi_k}(s', a')\, \pi(da'\mid s')p(ds'\mid s,a).
\]
This evaluation step corresponds to finding the fixed point of a contraction operator. It is performed iteratively, starting from a fixed function $Q$, typically $Q=0$, and repeatedly applying the Bellman update for $Q^{\pi_k}$.

\medskip
\noindent
\textbf{Policy improvement.} Define a new policy $\pi_{k+1}$ by choosing, at each state, an action maximizing the one-step lookahead based on $Q^{\pi_k}$:
\[
    \pi_{k+1}(s)
    \in
    \arg\max_{a\in\mathcal{A}}
   Q^{\pi_k}(s,a),
\]
essentially replacing the current action in state $s$ by one that has a higher value.
\medskip

This procedure is called \emph{policy iteration}.

Repeated policy evaluation and improvement steps eventually converge to an optimal policy. The corresponding optimal value function $V^\star$ then satisfies the \emph{Bellman optimality equation}
\begin{equation}
\label{eq:bellman-optimality}
    V^\star(s)
    =
    \max_{a\in\mathcal{A}}
    \left\{
        r(s,a)
        + \gamma \int_{\mathcal{S}} V^\star(s')\, p(ds'|s,a)
    \right\}.
\end{equation}

This viewpoint will be particularly useful in the next section: many RL methods can be interpreted as approximate forms of policy iteration, where either the policy evaluation step, the policy improvement step, or both, are carried out only approximately from data \citep{chan2022greedification}.

\medskip
\noindent
\textbf{Remark.} The above procedure assumes that the expectations appearing in the Bellman equations can be computed exactly. This is rarely the case in general continuous systems, where these integrals are typically intractable. In finite state-action spaces, however, they reduce to finite sums, yielding an algorithm with polynomial complexity in the size of the state and action spaces. When the state-action space is very large, for example when it arises from a fine discretization of a $d$-dimensional continuous state space, the computational complexity of dynamic programming grows exponentially with $d$. This phenomenon, known as the curse of dimensionality \citep{bellman1957dynamic}, makes exact solutions of the Bellman equation intractable except for low-dimensional problems. Practical control methods therefore rely on additional structure to obtain scalable solutions. In this tutorial, we present two such approaches. In adaptive control, a reference model and a structured controller parametrization lead to tractable adaptation laws with stability guarantees. In RL, policies and/or value functions are represented within a restricted function class, enabling optimization directly in a lower-dimensional parameter space.

\subsection{Divergence and convergence of the two fields}

Dynamic programming is the common cornerstone of both RL and optimal control. However, it is important to note that while RL is almost synonymous with seeking optimality, the control community historically views optimality as secondary to stability. In fact, the literature is replete with examples of control designs that are nominally optimal but lack robust stability---culminating in the celebrated shortest abstract in history \citep{Doyle1978}---and are thus often unusable in engineering applications.

In this section, we highlight cultural and methodological divergences between the fields that may help explain how the solution methods in the next section were developed.

\paragraph{Divergent paths: in-silico vs. in-vivo. }
Despite the shared dynamic programming origins, the two fields have diverged based on their primary {\em environments}.
On the one hand, RL has been developed largely within computer science and has traditionally focused on an agent interacting with an {\em in-silico} simulator. This sandbox environment supports an emphasis on algorithm design and permits millions of interactions with the simulator, often without stability or safety constraints. Consequently, algorithm convergence rates, sample efficiency, and the exploration-exploitation tradeoff (see Section~\ref{subsec:exploration}) are primary metrics of success.

On the other hand, control methods emerged from engineering applications and are consequently designed for deployment in the ``real world'' and {\em in-vivo}. Here, the ``agent'' is a controller for a physical plant. In this context, optimality is often a luxury. The priorities are stability, safety (synonymous with constraint satisfaction), robustness to unmodeled disturbances or uncertainties, and implementation concerns involving real-time computation, partial observations, parsimonious use of data, and so on.

\paragraph{Cultural epistemology: certificates vs. benchmarks. }
This divergence and the different environments have fostered distinct research cultures. In the control community, a ``valuable contribution'' typically requires formal certificates---mathematical guarantees of safety or stability for the real-world closed-loop system.

These are usually derived through pen-and-paper analysis (e.g., Lyapunov stability), and computational certificates have been widely embraced only relatively recently.
Conversely, the RL community prioritizes algorithmic certification, empirical performance, and scalability. Rather than tailoring solutions to domain-specific physics, RL has embraced the infamous {\em bitter lesson} \cite{sutton2019bitter}: the observation that general-purpose methods that leverage massive computation (e.g., deep RL) eventually outperform human-engineered features. Success in RL is thus typically defined by performance on standardized benchmarks or solving complex, high-dimensional tasks that defy classical approaches.

At the same time, modern control theory has deepened its understanding of robustness, safety, and performance guarantees for uncertain and nonlinear systems. Key developments, including robust adaptive control, Lyapunov-based analysis for learning systems, and learning-based model predictive control, have begun to address many of the limitations that once hindered adaptive systems in practice \citep{annaswamy2023adaptive,dean2020sample,hewing2020cautious}. Importantly, these developments show that learning and safety need not be at odds, and that adaptive behavior can be both data-driven and certifiable when properly designed \citep{ames2017control}.

\paragraph{Recent confluence.} These differences between RL and control are to be expected; they are logical consequences of their respective development in in-silico digital sandboxes and in-vivo physical systems. While their environments necessitated different priorities, the boundaries are now blurring.
Modern control has increasingly incorporated learning-based components, while RL has begun addressing safety and robustness in real-world systems. These trends suggest that these two histories are once again merging into a unified field of data-driven decision-making.

\section{Data-driven decision making}

Both RL and adaptive control address the same fundamental challenge: how to design controllers when the system dynamics are not fully known. Their primary difference lies in how uncertainty is modeled and handled, as well as in the types of guarantees that are sought.
A key bridge between the two fields is the literature on \emph{approximate dynamic programming} \citep{powell2007approximate, bertsekas1996neuro, lewis2009reinforcement}, which combines function approximation with dynamic programming principles.

This core technical section first reviews the model-reference adaptive control algorithm as well as the actor-critic family of RL algorithms.
We then present an original approach that combines both points of view in a low-level/high-level hybrid control strategy (Section~\ref{subsec:combining_RL_and_control}). This new approach shows that the two paradigms can indeed be complementary rather than opposed.

\subsection{Adaptive control}
\label{subsec:adaptive_control}

Adaptive control is the classical control-theoretic paradigm for updating a controller online in response to uncertain or changing dynamics while maintaining a stability-oriented design viewpoint. Among the many variants developed in this literature, MRAC \cite{goodwinMRAC1984,anderson1986stability,sastry1989adaptive} plays a particularly central role: one first specifies a stable reference model that encodes the desired closed-loop behavior, and then adjusts controller parameters in real time so that the plant tracks this reference behavior despite parametric uncertainty.

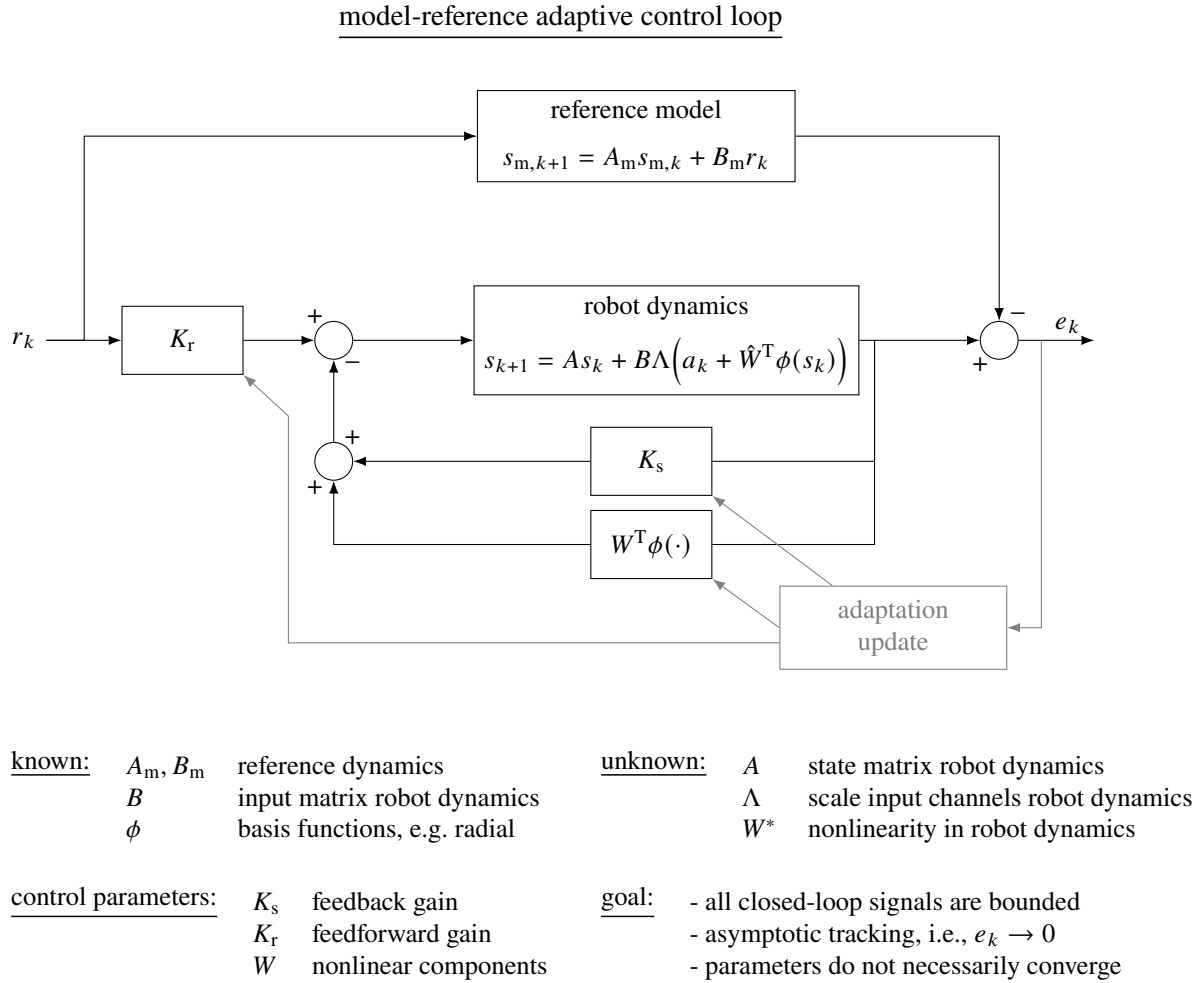
\begin{figure}

\centering
\usetikzlibrary{arrows.meta,positioning,calc}

\begin{tikzpicture}[
    >=Latex,
    auto,
    node distance=1.8cm and 1.8cm,
    block/.style={draw, rectangle, minimum height=1.2cm, minimum width=4.2cm, align=center},
    smallblock/.style={draw, rectangle, minimum height=0.9cm, minimum width=1.6cm, align=center},
    sum/.style={draw, circle, inner sep=0pt, minimum size=5mm},
    every node/.style={font=\small}
]

\node[left] (r) at (-6,0) {$r_k$};

\node[block] (ref) at (1.8,2.7)
{reference model\\[2mm]
$\displaystyle s_{\text{m},k+1}=A_\text{m} s_{\text{m},k}+B_\text{m} r_k$};

\node[block] (plant) at (2.2,0)
{robot dynamics\\[2mm]
$\displaystyle s_{k+1}=A s_k + B\Lambda\!\left(a_k+\hat{W}^{\mathrm{T}}\phi(s_k)\right)$};

\node[sum] (sum1) at (-2.2,0) {};
\node[sum] (sum2) at (-2.2,-1.6) {};
\node[sum] (sum3) at (6.6,0) {};

\node[smallblock] (Kr) at (-4.2,0) {$K_\text{r}$};
\node[smallblock] (Ks) at (2.0,-1.6) {$K_\text{s}$};
\node[smallblock] (Wphi) at (2.0,-2.7) {$W^{\mathrm{T}}\phi(\cdot)$};
\node[block, gray, minimum width=3.0cm, minimum height=1.1cm] (adapt) at (5.2,-3.8)
{adaptation\\update};

\node[right] at (7.2,0.2) {$e_k$};

\coordinate (rup) at (-5.5,0);

\draw[->] (r.east) -- (rup) |- (ref.west);

\draw[->] (r.east) -- (Kr.west);
\draw[->] (Kr.east) -- (sum1.west);

\draw[->] (sum1.east) -- (plant.west);

\draw[->] (plant.east) -- (sum3.west);

\draw[->] (ref.east) -| (sum3.north);

\draw[->] (sum3.east) -- ++(1.0,0);

\draw[->,gray] ([xshift=3mm]sum3.east) |- (adapt.east);

\coordinate (fbtap) at ($(plant.east)+(0.2,0)$);
\draw[-] (fbtap) |- (Ks.east);
\draw[-] (fbtap) |- (Wphi.east);

\draw[->] (Ks.west) -- ++(-1.5,0) |- (sum2.east);

\draw[->] (Wphi.west)  -| (sum2.south);

\draw[->] (sum2.north) -- (sum1.south);


\node at (-2.45,0.32) {$+$};
\node at (-1.95,-0.32) {$-$};

\node at (-1.95,-1.28) {$+$};
\node at (-2.45,-1.95) {$+$};

\node at (6.85,0.32) {$-$};
\node at (6.35,-0.32) {$+$};

\coordinate (tmp) at (-2.8,-1.);

\node[font=\normalsize] at (0.8,4.2) {\underline{model-reference adaptive control loop}};

\draw[->,gray] (adapt.west) -- (Wphi.south east);
\draw[->,gray] ([xshift=7mm]adapt.north west) -- (Ks.south east);
\draw[->,gray] (tmp) -- (Kr.south east);
\draw[-,gray] ([yshift=-2mm]adapt.west) -| (tmp);


\node[anchor=north west] (knownhdr) at (-6.6,-5.3) {\underline{known:}};
\node[anchor=north west, align=left] at ([xshift=2mm]knownhdr.north east) {%
\begin{tabular}[t]{@{}ll@{}}
$A_\text{m},B_\text{m}$ & reference dynamics \\
$B$       & input matrix robot dynamics \\
$\phi$    & basis functions, e.g.\ radial
\end{tabular}};

\node[anchor=north west] (unknownhdr) at (1.2,-5.3) {\underline{unknown:}};
\node[anchor=north west, align=left] at ([xshift=2mm]unknownhdr.north east) {%
\begin{tabular}[t]{@{}ll@{}}
$A$        & state matrix robot dynamics \\
$\Lambda$  & scale input channels robot dynamics \\
$W^*$  & nonlinearity in robot dynamics
\end{tabular}};

\node[anchor=north west] (controlhdr) at (-6.6,-7.1) {\underline{control parameters:}};
\node[anchor=north west, align=left] at ([xshift=2mm]controlhdr.north east) {%
\begin{tabular}[t]{@{}ll@{}}
$K_\text{s}$ & feedback gain \\
$K_\text{r}$ & feedforward gain \\
$W$   & nonlinear components
\end{tabular}};

\node[anchor=north west] (goalhdr) at (1.2,-7.1) {\underline{goal:}};
\node[anchor=north west, align=left] at ([xshift=2mm]goalhdr.north east) {%
\begin{tabular}[t]{@{}l@{}}
- all closed-loop signals are bounded \\
- asymptotic tracking, i.e.,\ $e_k \rightarrow 0$ \\
- parameters do not necessarily converge
\end{tabular}};

\end{tikzpicture}
\caption{Architecture of a model-reference adaptive controller. The adaptation loop is shown in gray. \label{fig:mrac}}
\end{figure}

Figure~\ref{fig:mrac} highlights the basic MRAC architecture.
The reference trajectory $r_k$ is filtered through a linear reference model with matrices $A_\mathrm{m}$ and $B_\mathrm{m}$, defining the desired response $s_{\text{m},k}$.
The physical system dynamics are represented by a nominal linear input channel together with unknown state dynamics $A$, unknown input scaling $\Lambda$, and a nonlinear term approximated through basis functions ${W^*}^\mathrm{T} \phi(\cdot)$.
The matrix $B$ (action directions) is assumed to be known. The actions are given by an adaptive controller that combines a feedforward gain $K_\text{r}$, a feedback gain
$K_\text{s}$, and a nonlinear compensation term $W^\mathrm{T} \phi(\cdot)$.
The tracking error $e_k \doteq s_k - s_{\mathrm{m}, k}$ between the realized system state and the reference-model state is then used in the adaptation law to update these parameters online, with the goal of keeping all closed-loop signals bounded and driving the tracking error to zero despite uncertainty in the system dynamics.

The key insight behind this architecture is the \emph{matching assumption}, namely that there exist ideal feedback gains $K_\text{s}^\ast$ and $K_\text{r}^\ast$ satisfying
\[
A-B\Lambda K_\text{s}^\ast=A_\text{m},
\qquad
B\Lambda K_\text{r}^\ast=B_\text{m}.
\]
These conditions imply that the closed-loop system can, in principle, reproduce the target dynamics specified by the reference model. They are the standard matching conditions in adaptive control \cite{narendra1989stable,sastry1989adaptive}; in RL terminology, they play a role similar to a realizability assumption \cite{mohriFoundations}, namely that an optimal policy exists within the considered model class. Since the plant matrices are unknown, the ideal gains $K_\text{s}^\ast$ and $K_\text{r}^\ast$ cannot be computed directly. Instead, the adaptive controller maintains estimates $K_\text{s}$, $K_\text{r}$, and $W$, and it is convenient to express them relative to the ideal parameters. Under the matching assumption, the combined system, controller, and reference model yield the tracking-error dynamics
\begin{equation}
e_{k+1}=A_\text{m}e_k+B\Lambda\Theta^\mathrm{T}\omega_k,\quad
\text{with}\quad
\Theta\doteq
\begin{pmatrix}
K_\text{s}^\mathrm{T}-{K_\text{s}^\ast}^\mathrm{T}\\
K_\text{r}^\mathrm{T}-{K_\text{r}^\ast}^\mathrm{T}\\
W- W^*
\end{pmatrix},
\quad\text{and}\quad
\omega_k\doteq
\begin{pmatrix}
s_k\\
r_k\\
\phi(s_k)
\end{pmatrix},
\label{eq:errordynamics}
\end{equation}
where $\Theta$ collects the parameter estimation errors. Denoting the dimensionality of $s_k$ by $d_s$, of $a_k$ by $d_a$, of $r_k$ by $d_r$, and of $\phi(s_k)$ by $d_\phi$, we have $e_k\in\mathbb{R}^{d_s}$, $\Theta\in\mathbb{R}^{(d_s+d_r+d_\phi)\times d_a}$, and $\omega_k\in\mathbb{R}^{d_s+d_r+d_\phi}$.

An adaptive controller now chooses parameter updates $\Theta_{k+1}$ such that the error dynamics~\eqref{eq:errordynamics} remain bounded and decay asymptotically. Given positive definite matrices $Q\in\mathbb{R}^{d_s\times d_s}$ and $\Gamma\in\mathbb{R}^{(d_s+d_r+d_\phi)\times(d_s+d_r+d_\phi)}$, the update law is typically derived from a Lyapunov function $U$ such as
\begin{equation*}
U(e,\Theta)=e^\mathrm{T}Pe+\operatorname{tr}(\Theta^\mathrm{T}\Gamma^{-1}\Theta\Lambda),
\end{equation*}
where $P$ solves the Lyapunov equation $A_\text{m}^\mathrm{T}PA_\text{m}-P=-Q$. The update law is chosen so that the Lyapunov function decreases along the trajectories of the adaptive system. For the case $d_a=\Lambda=1$, this update becomes
\begin{equation*}
\Theta_{k+1}=\Theta_k-\Gamma\omega_kB^\mathrm{T}Pe_k.
\end{equation*}
In continuous time, one can show that the corresponding decrease in the Lyapunov function is proportional to $-e^\top Qe$, implying that the tracking error decreases over time.

In discrete time, the same calculation introduces additional higher-order terms in the sampling time and step size $\Gamma$.

Neglecting these terms, the update yields the approximate decrease
\[
U(e_{k+1},\Theta_{k+1})-U(e_k,\Theta_k)
\approx
-e_k^\top Q e_k,
\]
suggesting that the tracking error contracts over time. Rigorous discrete-time analyses account for the higher-order terms explicitly and typically require sufficiently small adaptation gains, normalization, projection, or related modifications to guarantee boundedness and asymptotic tracking.

Even though the parameter update is expressed in terms of the parameter error $\Theta$, the ideal parameters $K^*_\text{s}$, $K^*_\text{r}$, and $W^*$ cancel when the update is translated to the parameter estimates and therefore do not need to be known.

The update law also admits a gradient-descent interpretation.
If one considers an instantaneous quadratic tracking cost such as $e_k^\mathrm{T} P e_k$, then the term $\omega_k B^\mathrm{T} P e_k$ can be viewed as a sensitivity-weighted descent direction with respect to the controller parameters.
Thus, at an intuitive level, MRAC updates the controller parameters to reduce the current tracking error in much the same way that online gradient descent would.
In the derivation above, though, the update is selected not only for error reduction, but also because it supports a dissipation inequality.

\paragraph{Illustration: Second-order system.}
\begin{figure}
    \includegraphics[width=\textwidth]{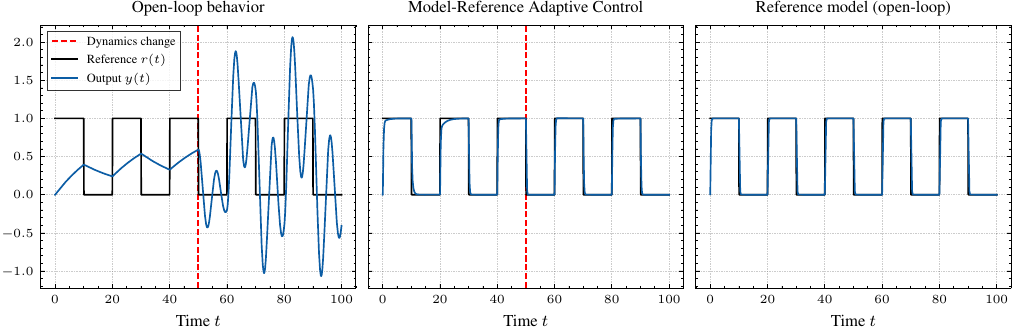}
    \caption{Illustration of model-reference adaptive control on a second-order system with a dynamics change. The goal is to make the output track the reference signal. (Left) Open-loop behavior of the system. The change in damping at $t = 50$ drastically changes the behavior. (Middle) MRAC automatically tunes the parameters of a feedback controller to accurately track the reference signal despite the dynamics change. (Right) Open-loop behavior of the reference model used by MRAC. \label{fig:mrac-example}}
\end{figure}
We now demonstrate the MRAC algorithm in a simple disturbance-free second-order dynamical system.
Figure~\ref{fig:mrac-example} illustrates such a system, in which the dynamics change at $t = 50$ from overdamped ($\zeta = 10$, ``high friction'') to underdamped ($\zeta = 0.1$, ``low friction'').
The natural angular frequency of the system remains constant at $\omega = 1$, and the sampling period for the zero-order hold discretization is $T = 0.01$.
Our goal is to make the output of this system (the displacement) follow the reference trajectory.
In the Half-Cheetah joint example \eqref{eq:sos}, this can be interpreted as $r(t)$ specifying the desired angle of the joint at time $t$, while $y(t) = \theta(t)$ is the actual angle.
If the reference signal is directly fed as an input to the system, the tracking error is very large (this is shown in the left plot).
To accurately track the reference signal, feedback is necessary.
Writing $\theta$ for the output, the general form of the input $u$ applied to the system is
\begin{equation*}
    u(t) = k_r r(t) + k_\theta \theta(t) + k_{\dot \theta} \dot \theta(t).
\end{equation*}
Plugging this into \eqref{eq:sos}, with $u = \tau/I$ and $\eta = 0$, the closed-loop joint dynamics are then given by
\begin{equation*}
    \ddot\theta + (2\zeta\omega - k_{\dot\theta}) \dot\theta + (\omega^2 - k_\theta) \theta = k_r r.
\end{equation*}
Thus, by choosing the controller parameters $k_r$, $k_\theta$, and $k_{\dot \theta}$ appropriately, the closed-loop system can follow any desired second-order dynamics.
We now use the MRAC update to tune these parameters.
For this, we first need to specify a reference model in which the open-loop behavior follows the reference trajectory closely.
In our case, we choose a second-order system with natural frequency $\omega = 10$ (fast response) and $\zeta = 1$ (critical damping).
The open-loop behavior of this system is illustrated on the right in Figure~\ref{fig:mrac-example}.
The center plot shows the trajectory of the real system under the MRAC controller with $\Gamma = 100I$.
Note that the controller parameters undergo a rapid change at $t=50$ to enable tracking the reference trajectory despite the change in dynamics.

\paragraph{Limitations.}

Despite its elegance, MRAC has well-known limitations.
These shortcomings motivated a substantial literature on robust and modified adaptive laws, including normalization, projection, dead zones, and composite adaptation, designed to improve robustness while preserving the core stability-oriented philosophy of adaptive control \citep{ioannou2012robust}.

\textbf{Take-home message. }
MRAC captures the canonical gray-box perspective of adaptive control emphasized throughout this paper.

The reference model specifies the desired evolution of each joint angle along a target gait trajectory, while the adaptive law adjusts controller gains online so that the realized angles continue to track these references even when the robot’s effective dynamics are altered by changes in link masses, joint friction, or other modeling mismatches. This decomposition is attractive because it preserves a familiar robotics control architecture---trajectory generation at a higher level and fast joint-level feedback underneath---while using adaptation to compensate for structured uncertainty at the actuator and joint-dynamics level.

\subsection{Adaptive control beyond MRAC}
\label{sec:beyond_mrac}

Beyond classical MRAC, adaptive control has developed along three major directions: (i) extension to increasingly complex system classes, including nonlinear, distributed-parameter, and delay systems; (ii) improved transient performance and robustness to modeling errors and disturbances; and (iii) understanding the role of excitation for learning and control.

A major step beyond linear MRAC was the introduction of adaptive backstepping \citep{krstic1995nonlinear}, a recursive Lyapunov-based methodology for nonlinear systems. Unlike classical MRAC, which primarily provides asymptotic guarantees, adaptive backstepping enabled systematic controller design for broad classes of nonlinear plants and led to techniques for improving transient performance under large parametric uncertainty. Subsequent developments extended these ideas to systems with unmodeled dynamics and external disturbances.

The same philosophy was later generalized to increasingly complex dynamical systems. In particular, backstepping-based methods were adapted to control partial differential equations (PDEs), yielding explicit controller constructions for distributed-parameter systems. When unknown plant parameters are estimated online and incorporated into these controllers, one obtains adaptive control schemes for PDEs \citep{smyshlyaev2010adaptive}. Related developments led to adaptive control methodologies for systems with unknown delays, where online estimation of delays can be incorporated into the controller design \citep{zhu2020delay}.

Another active line of research concerns robustness to non-parametric uncertainties such as disturbances and unmodeled dynamics. Classical mechanisms such as leakage and projection guarantee boundedness of the adaptive parameters but may introduce residual regulation errors that depend on unknown quantities. More recent approaches seek robustness guarantees that are independent of the magnitude of the parametric uncertainty. One recent example is deadzone-adapted disturbance suppression \citep{karafyllis2025robust}, which provides arbitrarily small regulation errors despite large parametric uncertainty.

Finally, adaptive control has long emphasized a key issue that is closely related to exploration in reinforcement learning (see Sec.~\ref{subsec:exploration}): the need for informative data. Accurate parameter estimation typically requires \emph{persistent excitation}, a condition ensuring that the observed trajectories contain sufficient information to identify unknown parameters. However, generating such excitation may conflict with the primary control objective of regulating the system. This tension is analogous to the exploration-exploitation tradeoff in reinforcement learning, where informative actions may temporarily degrade performance but improve future decision making. A notable achievement of adaptive control is that many adaptive controllers achieve strong regulation and tracking guarantees without requiring persistent excitation \citep{astrom2013adaptive,narendra1989stable}. Rather than aiming to identify the system completely, these methods learn only the information necessary to achieve the control objective. This perspective is also present in the RL literature, where exploration through optimism (see Sec.~\ref{subsec:exploration}) aims to collect only the amount of data needed to solve the reward-optimization problem.

\subsection{A standard RL approach: actor-critic algorithms}
\label{sec_rl_alg}

In this section we outline a canonical class of RL algorithms, called actor-critic. We will first present a more basic version, then discuss some of the specific implementation choices that matter when using neural networks (i.e., the deep RL setting).
We consider a setting where no white-box model nor high-fidelity simulator of the environment is available. Rather, we only have data, and potentially some knowledge of the problem.

The goal is to learn a parameterized policy $\pi_\theta:\mathcal{S}\to \Delta_{\mathcal{A}}$ that minimizes the objective introduced in Equation~\eqref{eq:main-objective}, or equivalently maximizes the expected discounted reward from the start states: $V(\pi) = \mathbb{E}[\sum_{t=0}^\infty \gamma^t r(s_t,a_t)]$.

To improve the policy, the agent must estimate which actions lead to high long-term rewards using the action-value function $Q^\pi(s,a)$ (Equation~\eqref{eq:Q-function}). Modern RL algorithms typically learn an approximation of this quantity and use it to guide policy improvement.

For example, $\pi_\theta$ could be a neural network with parameters $\theta$ that takes the state $s$ as input and outputs a multidimensional Gaussian distribution over actions.

A key result underpinning actor-critic methods is the policy gradient theorem \citep{sutton1999policy}. It shows that the gradient of the objective can be expressed without differentiating through the environment dynamics, relying only on the policy and its action-value function:

\begin{equation*}
    \nabla_\theta V(\pi)
    =
    \int d^{\pi_\theta}_\gamma(s)
    \int \pi_\theta(a|s)
    Q^{\pi_\theta}(s,a)
    \nabla \ln \pi_\theta(a|s)
    \, da\, ds.
\end{equation*}

The term $d^\pi_\gamma(s)$ corresponds to the discounted state-visitation distribution under policy $\pi$.\footnote{Note that the definition of $d^\pi_\gamma(s)$ is slightly different between the episodic and continuing settings in RL. In the continuing setting, $d^\pi_\gamma(s)$ corresponds to the stationary distribution for the policy $\pi$, with no dependence on $\gamma$. Otherwise, the rest of the gradient remains the same. Due to this similarity, algorithms designed for the episodic setting can often be applied heuristically to the continuing setting, although they should account for differences in state weighting. To clarify terminology, the \emph{episodic} setting corresponds to \emph{batch control} problems and the \emph{continuing} setting corresponds to the \emph{continuous} setting in control. This terminology should not be confused with the fact that in RL the continuous setting usually means the state and/or action are continuous rather than discrete.}

The policy gradient theorem suggests a simple strategy: sample states and actions from the current policy and weight the policy update by the corresponding action-value.

Obtaining unbiased samples of this gradient, as in algorithms such as REINFORCE \citep{williams1992simple}, is typically not sample efficient; on-policy algorithms such as proximal policy optimization obtain nearly unbiased samples \citep{schulman2017proximal}.
Instead, many algorithms estimate the action-value function to obtain a biased approximation of the gradient from a state $s$.

Actor-critic methods replace the unknown quantity $Q^{\pi_\theta}(s,a)$ appearing in the policy gradient theorem with a learned approximation $q_w(s,a)$.
The policy $\pi_\theta$ is referred to as the \emph{actor}, while the learned action-value approximation $q_w$ is called the \emph{critic}. The actor chooses actions and the critic evaluates their long-term consequences. Conceptually, this mechanism can be viewed as a form of approximate policy iteration. The critic approximates the value for the current policy (actor)
and the actor uses these approximations to improve the policy.

Specifically, the algorithms use the gradient approximation at state $s$
\begin{equation}\label{eq_ac_policy}
    \Delta\theta
    \doteq
    q_w(s,a)\nabla \ln \pi_\theta(a|s)
    \qquad
    \text{using }
    a\sim\pi(\cdot|s).
\end{equation}
The critic $q_w$ gives an estimate of the expected return (value) from the current policy, instead of having to run the current policy for multiple steps (a rollout) to get an unbiased sample of the return.
As the policy $\pi_{\theta_t}$ changes with parameters $\theta_t$ on time step $t$, $q_{w_t}$ is also updated to track an estimate of the action-values for $\pi_{\theta_t}$.

\paragraph{Illustration on the Half-Cheetah. } Let us consider more precisely how this algorithm behaves while interacting with our running example, Half-Cheetah. The agent starts in an initial state $s_0$ and samples an action $a_0$ from its stochastic policy $\pi_{\theta_0}$, which is typically set to be a Gaussian policy. The action space is bounded, so a tanh parameterization is used to respect the action bounds: $\pi_\theta(a|s) = \tanh(g_\theta(s,\epsilon))$, where $g_\theta(s,\epsilon) = \mu_\theta(s) + \epsilon\sigma_\theta(s)$ and $\epsilon \sim \mathcal{N}(0,I)$ is a vector of normal variables, one for each action dimension; $\mu_\theta(s)$ and $\sigma_\theta(s)$ are the outputs of the neural network. If there are multiple actions, it is typical to use independent Gaussians for each action. Note that, despite this choice, the joint utility of actions is still considered, because the critic evaluates the joint utility of all actions. The primary limitation of using independent Gaussians is that they do not allow for correlations when sampling actions, though the mean can still converge to the optimal multivariate action in each state.

After taking action $a_0$, the environment transitions to state $s_1$ and the agent observes reward $r_1$. The agent adds this transition $(s_0, a_0, r_1, s_1)$ to its replay buffer. It is typical to wait for the replay buffer to fill for some number of steps before updating, meaning that the agent takes random actions during the first few steps. If previously collected offline data are available, the agent can initialize its replay buffer with these historical data and omit this initial exploration phase. In our Half-Cheetah experiment, for example, this exploration phase lasts 100 steps, and the maximum size of the replay buffer $\mathcal{D}$ is 100,000 samples.

Once the buffer has sufficient samples, both the actor and critic are updated at each step using sampled minibatches $\mathcal{B} \subset \mathcal{D}$ from the replay buffer $\mathcal{D}$. For Half-Cheetah, a typical minibatch size is $b=256$, meaning that $\mathcal{B}$ contains 256 tuples $(s,a,r,s')$ randomly selected from $\mathcal{D}$. The action-value function (critic) $q_w$ is updated using a TD update on these tuples:
\begin{equation}
    \Delta w \doteq \frac{1}{b} \sum_{(s,a,r, s') \in \mathcal{B}} (r +\gamma q_w(s', a' \sim \pi_\theta(\cdot | s')) - q_w(s,a)) \nabla q_w(s,a)
    .
\end{equation}

The policy is updated using the same minibatch, though only the state is used:
\begin{equation}
    \Delta \theta \doteq \frac{1}{b} \sum_{s \in \mathcal{B}, a\sim \pi_\theta(\cdot | s)} (q_w(s,a) - v(s)) \nabla \ln\pi_\theta(a|s)
    .
\end{equation}

For this \emph{log-likelihood} (score function) form of the gradient, it is typical to include an estimate $v(s)$ of the value function $\mathbb{E}_{a \sim \pi_\theta(\cdot|s)}[q_w(s,a)]$.
This centered term, $q_w(s,a) - v(s)$, reflects the \emph{advantage} of one action over another in the given state. It is often used as a heuristic to provide a
more effective gradient update because it is negative for low-valued actions and positive for high-valued actions. When the policy distribution can be reparameterized---as it can be for the Gaussian---and the critic $q_w$ is differentiable with respect to the input actions, a typically more effective alternative is the \emph{reparameterized} gradient update obtained by differentiating $q_w(s,\tanh(g_\theta(s,\epsilon)))$ with respect to $\theta$ \citep{haarnoja2018soft}. These minibatch stochastic gradient updates for the critic and actor are applied at each step---potentially multiple times per step---using standard machine-learning optimizers such as Adam \citep{kingma2014adam}, which uses vector step sizes and momentum. The replay buffer is a sliding window, with the oldest samples dropped as new interactions are added.

The above update strategy forms the basis of many actor-critic updates in deep RL, albeit with a variety of implementation heuristics introduced to improve performance. Soft actor-critic (SAC) \citep{haarnoja2018soft} uses entropy regularization to maintain stochasticity in the policy, learns two critics to reduce overestimation in the critic's bootstrap target, uses a slowly changing target to promote stability, and applies the reparameterized gradient update to the actor. The approach can also be extended to learning deterministic policies with an algorithm called DDPG \cite{silver2014deterministic} and a corresponding variant, TD3 \citep{fujimoto2018addressing}, which similarly adds a variety of implementation heuristics to improve performance.

\textbf{Take-home message. } Actor-critic methods exploit differentiable parametrized representations of the policy and action-value functions. At each update, an approximate gradient of the optimization objective is computed directly using data from previous rollouts stored in the replay buffer. Concretely, these data are used to update the representations of the actor $\pi_\theta$ and the critic $q_w$. Because these algorithms do not need to estimate a model of the dynamics, they are called \emph{model-free}. They are designed to learn optimal behavior from scratch in terms of the cumulative-reward objective, without using reference trajectories or demonstrations.

\paragraph{Limitations. } These algorithms lack some critical components of the RL problem setting: directed exploration, handling partial observability, and, in many cases, even soundness. We discuss exploration and partial observability in the next section but focus on soundness here. The primary issue is that these algorithms are largely based on heuristics. Their updates use biased gradients,

and in some cases these are known to diverge \citep{tsitsiklis1996analysis} or converge to suboptimal solutions \citep{thomas2014bias,graves2023off,patterson2022generalized}. Typically, the state weighting $d^\pi_\gamma(s)$ is ignored, and updates are instead drawn randomly from the replay buffer, which has a different implicit distribution. This mismatch can cause suboptimal solutions, though there is some theoretical and empirical evidence that, with sufficiently high-capacity (large) neural networks, it does not cause issues \citep{graves2023off}. Another key issue arises when moving from linear function approximation, for which we have more theoretical guarantees \citep{zhang2020provably,karmakar2018two,chen2025convergence}, to nonlinear function approximation with neural networks in deep RL. For the critic, the TD update is known to diverge under nonlinear function approximation \citep{tsitsiklis1996analysis}. A variety of gradient TD approaches have been proposed to remedy this problem \citep{patterson2022generalized}, along with some theoretical results for actor-critic methods with neural networks \citep{gaur2024closing}. These more principled approaches, however, have not yet been widely adopted, and commonly used standard algorithms such as SAC and TD3 have no such guarantees.
These issues are recognized, and progress is being made toward sounder and more robust variants of actor-critic methods that use neural networks. Today, however, these algorithms must still be used with caution and have not reached the same level of maturity as RL algorithms for tabular or linear settings or as algorithms in control.

Another limitation of these algorithms is that they are designed to find the optimal policy without considering performance during learning. Following the direction of the gradient to optimize the objective does not ensure good transient performance. Relatedly, stability and constraints are not encoded in this objective, and any stability or safety properties must be encoded in the costs themselves; even then, they could be violated during learning. At the same time, it is worth noting that finding the optimal policy is not strictly at odds with good transient behavior. Acting nearly greedily does push the algorithm toward reducing cost sooner. The agent adjusts based on feedback from its interaction with the environment and will reduce costs, at least locally. The RL literature includes many works that report transient behavior and rank algorithms accordingly (see \cite{patterson2024empirical}), despite the fact that the objective itself does not optimize for transient behavior. Finally, there is a growing literature on safe RL that provides guarantees on constraint satisfaction during learning \citep{sukhija2023gosafeopt,wang2023enforcing,hsu2023safety} or uses offline data to learn control barrier functions \citep{tayal2026vocbf}.

\subsection{Combining RL and adaptive control}
\label{subsec:combining_RL_and_control}
\begin{figure}
    \includegraphics[width=\textwidth]{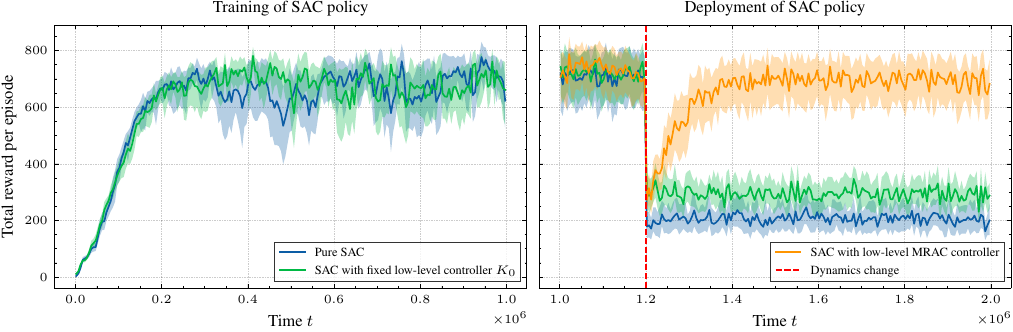}
    \caption{SAC in the Half-Cheetah environment.
    The left plot shows the first $10^6$ time steps, in which an SAC policy is trained.
    The right plot shows the second $10^6$ time steps, in which the learned SAC policy is deployed.
    After 200,000 steps of deployment, the friction in the Half-Cheetah joints changes, causing the non-adaptive policies to struggle.
    The orange line shows our proposed SAC+MRAC solution, in which a low-level adaptive controller is combined with a fixed SAC policy.
    The low-level MRAC controller aims to restore the nominal low-level dynamics encountered in the first 200,000 steps of deployment of the SAC-$K_0$ policy, on the basis of which a linear reference model is learned.
    }
    \label{fig:sac}
\end{figure}
As described above, modern RL algorithms like SAC usually focus on learning a fixed policy that achieves high performance in a static environment.
In this section, we examine how SAC can be combined with a low-level MRAC controller to adapt this fixed policy to changes in the environment dynamics.
Training a policy in a static environment is straightforward, and Figure~\ref{fig:sac} (left, blue line) shows that SAC manages to make the Half-Cheetah run after about 200,000 samples from the environment, after which policy performance plateaus.
If we deploy this fixed policy in an environment whose dynamics change, the performance will drop.
This is illustrated on the right in Figure~\ref{fig:sac} (blue line), where, at $t = 1.2 \times 10^6$, the damping coefficients of all six Half-Cheetah joints are reduced by a factor of 2, as might occur after lubrication.
The performance of SAC drops drastically because the applied torques do not account for the reduced friction of the joints.
What has not changed, however, is the type of gait that the Half-Cheetah should follow.
In other words, if the policy were to choose a set of desired joint angles instead of joint torques, then these angles would still be applicable after the dynamics change.

Following desired angles is a reference tracking problem, as in Section~\ref{subsec:adaptive_control}.

In this case, the reference signal $r(t)$ is the output of the SAC policy, and we apply a low-level controller with parameters $k_r$, $k_\theta$, and $k_{\dot\theta}$ to compute the action $a(t)$ at time $t$:
\begin{equation*}
    a(t) = k_r r(t) + k_\theta \theta(t) + k_{\dot \theta} \dot \theta(t).
\end{equation*}
Modeling the joints again as second-order systems, the closed-loop dynamics are
\begin{equation}\label{eq:ddth}
    \ddot\theta + (2\zeta\omega - k_{\dot\theta}) \dot\theta + (\omega^2 - k_\theta) \theta = k_r r.
\end{equation}
A reasonable start would be to set $k_{\dot\theta} = 0$ and $k_r = -k_\theta \gg \omega^2$.
Then, the angles approximately follow
\begin{equation*}
    \ddot\theta \approx -2\zeta\omega\dot\theta - k_r(\theta - r),
\end{equation*}
meaning that the angle converges to $r$ at a speed depending on $k_r$, $\zeta$, and $\omega$.
Since the Half-Cheetah system requires actions $a$ to lie in $[-1, 1]$, we cannot choose $k_r$ arbitrarily high.

In Figure~\ref{fig:sac}, we show how SAC behaves when combined with a fixed low-level controller $K_0$, in which $k_{\dot\theta} = 0$ and $k_r = -k_\theta = 1$.
During training and deployment in the original environment, the performance is on par with pure SAC.
After the dynamics change, the performance is slightly better than pure SAC, but not by much.
Just like in the second-order system discussed in Section~\ref{subsec:adaptive_control}, a change in dynamics requires a change in control parameters.
And just like in Section~\ref{subsec:adaptive_control}, we will use the MRAC algorithm to tune the parameters automatically.

Since the SAC-$K_0$ combination works well under the original dynamics, we can use this system to create a reference model.
Specifically, we collect a dataset of 200,000 steps of interaction, recording $(r_k, \theta_k, \dot\theta_k, \theta_{k + 1}, \dot\theta_{k + 1})$ at every discrete time $k$, for each of the six joints.
We then find a least-squares fit for the matrices $A$ and $B$ such that
\begin{equation*}
   \begin{bmatrix}\theta\\\dot\theta\end{bmatrix}_{k + 1} \approx A \begin{bmatrix}\theta\\\dot\theta\end{bmatrix}_{k} + B r_k
\end{equation*}
These matrices specify our reference model, and $B$ is also used as the `real' input matrix in the MRAC update.
Note that if the original system dynamics obey Equation~\eqref{eq:ddth} and the change in dynamics affects only the damping coefficient $\zeta$, it is only necessary to adjust the coefficient $k_{\dot\theta}$ to restore the original dynamics.
We thus set the components of $\Gamma$ corresponding to $k_r$ and $k_\theta$ to zero and the component corresponding to $k_{\dot\theta}$ to $10^{-7}$.
Such a small step size is necessary to prevent instabilities that arise due to the neglected nonlinear nature of the system.
Similarly, we clamp $k_{\dot\theta}$ to the interval $[-0.3, 0.3]$.
Finally, the MRAC update depends on the error term $e = s - s_{\mathrm{m}}$, where the model state $s_{\mathrm{m}}$ follows the reference dynamics.
In our case, the reference dynamics are linear and given by the $A$ and $B$ matrices described above.
As the real Half-Cheetah system exhibits highly nonlinear contact dynamics that are impossible to realize in the reference model, the error term becomes very large and uninformative.
We thus replace the error $e$ by the one-step prediction error, effectively resetting $s_\mathrm{m}$ to $s$ at every time step.
This is a significant change to the MRAC update derived in Section~\ref{subsec:adaptive_control}, which does not preserve theoretical guarantees that apply in the linear setting.
We do not include a nonlinear $W^\text{T}\phi$ term in the controller.
The final behavior is shown in Figure~\ref{fig:sac} (right, orange line).
It can be seen that the adaptive controller restores the low-level dynamics to the nominal dynamics, which enables the SAC policy to effectively control the robot.
Our SAC implementation uses the hyperparameters mentioned in Section~\ref{sec_rl_alg}.
The code is published at \url{https://github.com/onnoeberhard/forlac-tutorial}.

\textit{Remark:}
An alternative approach would be to continue updating the SAC policy after the dynamics change. Adapting RL algorithms to nonstationary environments is an active area of research, involving questions such as how to detect changes in the environment, how to incorporate new experience while retaining useful past knowledge, and how to adapt sufficiently quickly to maintain performance. The goal of our example is not to suggest that a hybrid RL--adaptive control architecture is the only viable solution to such problems. Rather, it provides a simple and illustrative example of how the two methodologies can be combined. In this setting, the RL component learns a complex locomotion strategy, while the adaptive controller compensates online for changes in the system dynamics. The resulting architecture highlights the complementary strengths of the two fields and serves as a concrete illustration of the broader theme of this tutorial.

\section{Additional considerations}

In this section, we review and contrast a few additional key aspects of RL and control algorithms: the use of models, exploration, and the handling of partial observability.

\subsection{Model-based versus model-free algorithms}

Control has historically been mostly applied to well-known physical systems, and the predominant paradigm is that of a known gray-box model with parametric uncertainty. Access to a dynamics model, even only a gray-box model, is central in control theory and drives both algorithm design and the derivation of guarantees through Lyapunov analysis. Where such a first-principles model is a reasonable reflection of the deployment environment, this model can be used for powerful algorithms (e.g., model predictive control). It also facilitates incorporating constraints, and it can be instrumental in the analysis while providing some interpretability of the controller.

When no gray-box model of the environment is available, both fields have developed parallel approaches. While it is most common in RL to assume minimal prior knowledge of the system, we still make a distinction between \emph{model-based} and \emph{model-free} algorithms.
In \emph{model-based RL}, a model of the system dynamics is learned from data and subsequently used for planning, typically via dynamic programming or decision-time optimization \citep{sutton1991dyna}. The model can be used for multi-step rollouts (planning) or for improving value and policy estimates.
In contrast, \emph{model-free RL} algorithms aim to directly estimate value functions, policies, or their gradients from data, without explicitly constructing a model of the environment, similarly to actor-critic methods presented in the previous section.
In control, similar paradigms are known as \emph{direct} or \emph{indirect} control. The process of learning a model offline through data collection is known as \emph{system identification}, which can be compared to the model-based RL approach. An important difference, though, is that model-based RL algorithms usually do not attempt to estimate a full model but rather only what is useful for the desired objective, very much in line with the recent developments in adaptive control discussed in Sec.~\ref{sec:beyond_mrac}.

The distinction between model-based and model-free methods in RL, however, is not always sharp. In particular, the replay buffer used in many model-free methods can be interpreted as a form of nonparametric model of the dynamics \citep{pan2018organizing,van2019use}. From this perspective, updating value functions using stored transitions resembles planning with an implicit model defined by the data. Moreover, in complex environments where accurate parametric models are difficult to learn, replay buffers can be competitive with, or even outperform, learned parametric models \citep{aminmansour2024mitigating}.
There have been continued efforts to leverage learned models more effectively in RL, including methods that combine model learning with value estimation and policy optimization \citep{hafner2025training,lawrence2025mpcritic,hansen2023td}. Nonetheless, the dominant paradigm remains to rely directly on data through replay buffers for repeated updates. Conversely, modern control actively reflects on using data together with model learning \citep{dorfler2023data}.

It is worth noting that, although RL is conceptually formulated as learning through interaction with an unknown environment, in practice it is often applied in settings where a simulator is available \citep{bellemare2020autonomous,degrave2022magnetic,da2025survey}. In such cases, the simulator effectively acts as a generative model that can be queried to produce data. However, the simulator is typically treated as part of the environment rather than explicitly exploited for planning, and many widely used algorithms remain model-free even in this setting.

\subsection{Exploration and finite-time performance}
\label{subsec:exploration}

Most RL algorithms are based on data collection using a current policy. This poses the problem of \emph{coverage}: how can we guarantee that the data is sufficient to improve the policy?
This issue already appears in classical convergence results for tabular RL algorithms such as Q-learning and SARSA \citep{tsitsiklis1994asynchronous}. These guarantees typically rely on a persistent exploration condition: all state-action pairs must be visited infinitely often. Simple strategies such as $\epsilon$-greedy exploration (a form of uniform random persistent excitation) enforce this condition and ensure asymptotic convergence under suitable assumptions \citep{tsitsiklis1994asynchronous}. However, they do not control how efficiently information is gathered and may require a large number of samples to achieve good performance. In this section, we discuss how regret minimization formalizes the problem of minimizing data usage in RL and describe more efficient exploration algorithms.

\paragraph{Finite-time performance and regret.}
From a statistical perspective, asymptotic convergence is not sufficient: one is interested in the performance of the algorithm after a finite number of interactions with the system, so we need to account for such data collection steps.

The notion of \emph{regret} is central in RL theory and captures the performance of the learner during the transient regime compared to that of the optimal policy:
\begin{equation}
    \text{Regret}(T) = \sum_{t=1}^T \big( V(\pi^*) - V(\pi_t) \big).
\end{equation}
Minimizing regret requires balancing exploration and exploitation: exploring uncertain actions to improve future decisions, while exploiting current knowledge to achieve low immediate cost. This exploration-exploitation trade-off has been a central topic of research at the intersection of RL, online learning, and statistical learning theory over the past two decades.

\paragraph{Optimism and efficient exploration.}
A key principle underlying many theoretically efficient RL algorithms is \emph{optimism in the face of uncertainty} \citep{Auer2002UCB} (or OFU). The idea is to construct estimates of the value function together with confidence intervals on these estimates. Then, an optimistic proxy of the value function is constructed by combining raw estimates and uncertainty in a way that favors poorly explored state-action pairs, thereby encouraging the agent to gather information where it is most needed.
This mechanism, which was discovered independently in the control and statistics communities \citep{KuBe82,LaiRobbins85}, leads to algorithms with provable regret guarantees \cite{jaksch10ucrl}.

\paragraph{Exploration in deep RL.}
In large or continuous state spaces, the tabular exploration strategies described above cannot be applied directly. Modern deep RL algorithms instead rely on a variety of heuristic mechanisms to encourage exploration while learning value functions or policies represented by neural networks. In discrete domains, a prominent example is the use of Monte-Carlo tree search with optimism-based exploration bonuses, as in the upper confidence tree algorithm \cite{kocsis2006uct}. This approach was a key ingredient in the AlphaGo and AlphaZero systems \cite{silver2016alphago,silver2017alphazero}, where neural networks approximate value and policy functions while tree search provides targeted exploration of promising actions.

Other approaches attempt to estimate uncertainty directly in the learned value function, for example through bootstrapped ensembles \citep{osband2016deep}, posterior sampling \citep{Strens00,osband2017posterior,muehlebach2025sample}, or intrinsic motivation signals based on prediction error or state novelty \citep{martius2013information,bellemare2016unifying}. While a large body of work has been devoted to extending these ideas beyond the tabular and linear settings \citep{LaSze25}, there is little, if any, work aimed at bridging the gap between theoretically sound algorithmic principles and heuristics used in practice.

\paragraph{Exploration and excitation in continuous control.}
In continuous control settings, exploration is often implemented by randomizing the policy during data collection, for instance by injecting noise into the control actions or by explicitly encouraging high-entropy policies. From a control perspective, this idea is closely related to the notion of \emph{persistent excitation}, which ensures that the system is sufficiently stimulated to identify unknown parameters or dynamics. Recent work in RL has begun to revisit this connection \citep{eberhard2023pink}, emphasizing that informative excitation of the system can play a crucial role in enabling efficient learning of dynamics and value functions in continuous environments \citep{muehlebach2025sample}. However, designing exploration strategies that simultaneously guarantee stability, safety, and efficient learning remains an open problem at the intersection of RL and adaptive control.

\subsection{Partial observability}
\label{subsec:partial_obs}

Many techniques from both RL and control theory rely heavily on knowledge of the system \emph{state}.
Environments where the state is revealed at all times are called ``fully observable.''
This knowledge enables many techniques that would otherwise not be possible.
The reason is that the state is \emph{Markovian}: for predicting the future, there is no relevant information in the history that is not contained in the current state. Most realistic environments, however, are \emph{partially observable}, meaning that only observations derived from the state are revealed.
For example, in the Half-Cheetah system, the observations at time $t$ might include only the angular velocities of the joints, not the corresponding angles.
Thus, past observations have to be used to infer the full state; in this case, the angles can be obtained by integrating the angular velocities.

There are two fundamental ways to approach partially observable environments: estimating \emph{belief states} or directly summarizing histories.
In control theory, it is more typical to estimate the belief state from the stream of observations.
If the observations $y$ are randomly sampled from an emission distribution $p(y \mid x)$ conditioned on the state $x$, then this can be done by Bayesian filtering.
In particular, the goal is to compute the belief distribution over the state $x_t$ given the history of observations $y_0, \dots, y_t$.
This can be done recursively, as
\begin{equation*}
    p(x_t \mid y_0, \dots, y_t) = \frac{p(x_t \mid y_0, \dots, y_{t-1})p(y_t \mid x_t)}{\int p(x'_t \mid y_0, \dots, y_{t-1})p(y_t \mid x'_t)\,dx'_t},
\end{equation*}
and
\begin{equation*}
    p(x_t \mid y_0, \dots, y_{t-1}) = \int p(x_{t-1}' \mid y_0, \dots, y_{t-1}) p(x_t \mid x_{t-1}')\,dx_{t-1}'.
\end{equation*}
In practice, these integrals are often intractable to compute, but in the special case of a Gauss-Markov model, where the transition and emission probabilities are linear-Gaussian, these integrals can be computed analytically.
In this case, this recursive computation results in the celebrated \emph{Kalman filter}.
Furthermore, in a linear-quadratic-Gaussian dynamical system, it is provably optimal to choose actions based only on the mean of the belief distribution.
This strategy is called \emph{certainty equivalence control}; in nonlinear systems, such a strategy is generally suboptimal.
In many systems, the degree of uncertainty in the belief distribution matters, since it may be advantageous to execute ``information-gathering'' actions to reduce this uncertainty and enable a more informed decision about the optimal strategy.

Explicitly tracking the belief state is not very common in RL.
There is a large literature in RL on partially observable Markov decision processes (POMDPs), which involves estimating belief states, but much of this work assumes discrete states, even when applied to continuous settings like those in robotics \citep{lauri2022partially}.
In general, the belief distribution is not only difficult to compute, but also difficult to handle.
Further, it is not strictly necessary for good control:
the history contains the same information as the belief distribution.

Thus, in RL, it is more common to summarize the history directly, either by using fixed memory mechanisms such as frame stacking \citep{mnih2015human} and exponential moving averages \citep{janjua2024gvfs,eberhard-2025-partially}, or by learning a recurrent neural network memory end-to-end, jointly with the policy \citep{kapturowski2018recurrent,elelimy2024real}.
These approaches often work well in practice, despite the fact that the general partially observable RL setting is provably intractable, both computationally and statistically \citep{papadimitriou,littman1994memoryless,krishnamurthy2016pac}.
To explain this discrepancy, several recent works have identified classes of partially observable environments in which these worst-case results do not apply \citep{liu2022when,golowich2023planning,eberhard-2026-commit,zhao-2026-why}.
Extending these classes is an important open problem in RL theory.

\section{Conclusions and open problems}
\label{sec:conclusion}

RL and optimal control share a common foundation in dynamic programming but have developed distinct perspectives shaped by their respective domains. RL emphasizes data-driven optimization and scalability, while control focuses on stability, robustness, and formal guarantees.

In this tutorial, we presented a unified view of the two fields by comparing their problem formulations, learning paradigms, and methodological tools. We discussed how adaptation arises through RL and adaptive control, and clarified the role of models and data in model-based and model-free approaches.

Our main takeaway is that many differences between the two fields stem from differing assumptions and priorities rather than fundamentally different problems. As both communities increasingly address complex, real-world systems, their methods are converging toward a shared framework for data-driven decision making that balances performance with reliability.

\textbf{Open Problems. } In addition to the specific open problems discussed in the sections above, we mention a few research areas where we believe that synergies between the RL and control communities could be particularly valuable. One such area is long-horizon planning in robotics, where agents must reason over extended time scales while still satisfying short-term stability and safety requirements. Another is the design of hierarchical architectures that integrate high-level decision making with low-level execution, allowing abstract plans to be translated into reliable motor commands or process-level control actions. A closely related challenge is to understand how learning should be distributed across these levels: low-level controllers may adapt to uncertain dynamics and disturbances, while high-level policies may learn task structure, constraints, and strategies for exploration. This also raises the question of how experience can be consolidated across time scales, for instance through model-based rollouts, replay, or ``dreaming'' mechanisms that allow an agent to refine its behavior without direct interaction with the physical system. Further open problems include partial observability, where controllers must act based on incomplete or noisy measurements, and robust reinforcement learning, where policies must retain reliable performance under distribution shifts, modeling errors, and adversarial or unexpected disturbances. Importantly, for real-world deployment, safety (i.e., hard constraint satisfaction) is non-negotiable, which poses a formidable challenge for any policy, whether derived using RL or classical control.
More broadly, many real-world systems require a careful interplay between feedforward and feedback mechanisms: feedforward components can anticipate desired behavior or compensate for predictable effects, while feedback controllers remain essential for robustness to uncertainty, disturbances, and modeling errors. Progress on these questions will further consolidate the impact of RL and control theory on complex technological, social, and physical systems, ranging from robotics and industrial processes to foundation models and agentic artificial intelligence.

\section*{Acknowledgments}
C. Vernade thanks the German Research Foundation (DFG) for its support through project 468806714 of the Emmy Noether Programme and under Germany's Excellence Strategy---EXC number 2064/1, project number 390727645. C. Vernade also gratefully acknowledges funding from the European Union (ERC, ConSequentIAL, 101165883). Views and opinions expressed are, however, those of the author(s) only and do not necessarily reflect those of the European Union or the European Research Council. Neither the European Union nor the granting authority can be held responsible for them. Claire Vernade and Onno Eberhard thank the International Max Planck Research School for Intelligent Systems for its support. Martha White would like to thank the Natural Sciences and Engineering Research Council of Canada (NSERC), the Canada Research Chair program, the Canada CIFAR AI Chair Program, and the Alberta Machine Intelligence Institute for research support. Csaba Szepesvári gratefully acknowledges funding from the Canada CIFAR AI Chairs Program, Amii, and NSERC. Michael Muehlebach thanks the German Research Foundation for its support.

\bibliographystyle{abbrvnat}
\bibliography{main}

@article{elelimy2024real,
  title={Real-time recurrent learning using trace units in reinforcement learning},
  author={Elelimy, Esraa and White, Adam and Bowling, Michael and White, Martha},
  journal={Advances in Neural Information Processing Systems},
  volume={37},
  pages={17006--17043},
  year={2024}
}

@inproceedings{kapturowski2018recurrent,
  title={Recurrent experience replay in distributed reinforcement learning},
  author={Kapturowski, Steven and Ostrovski, Georg and Quan, John and Munos, Remi and Dabney, Will},
  booktitle={International conference on learning representations},
  year={2018}
}

@article{janjua2024gvfs,
  title={GVFs in the real world: making predictions online for water treatment},
  author={Janjua, Muhammad Kamran and Shah, Haseeb and White, Martha and Miahi, Erfan and Machado, Marlos C and White, Adam},
  journal={Machine Learning},
  volume={113},
  number={8},
  pages={5151--5181},
  year={2024},
  publisher={Springer}
}

@article{lauri2022partially,
  title={Partially observable markov decision processes in robotics: A survey},
  author={Lauri, Mikko and Hsu, David and Pajarinen, Joni},
  journal={IEEE Transactions on Robotics},
  volume={39},
  number={1},
  pages={21--40},
  year={2022},
  publisher={IEEE}
}

@article{tsitsiklis1994asynchronous,
  title={Asynchronous stochastic approximation and Q-learning},
  author={Tsitsiklis, John N},
  journal={Machine learning},
  volume={16},
  number={3},
  pages={185--202},
  year={1994},
  publisher={Springer}
}

@article{tayal2026vocbf,
	title = {V-{OCBF}: {Learning} {Safety} {Filters} from {Offline} {Data} via {Value}-{Guided} {Offline} {Control} {Barrier} {Functions}},
      journal={Transactions on Machine Learning},
	author = {Tayal, Mumuksh and Tayal, Manan and Singh, Aditya and Kolathaya, Shishir and Prakash, Ravi},
	year = {2026}
}

@article{sukhija2023gosafeopt,
	title = {{GoSafeOpt}: {Scalable} {Safe} {Exploration} for {Global} {Optimization} of {Dynamical} {Systems}},
	volume = {320},
	journal = {Artificial Intelligence},
	author = {Sukhija, Bhavya and Turchetta, Matteo and Lindner, David and Krause, Andreas and Trimpe, Sebastian and Baumann, Dominik},
	year = {2023}
}

@inproceedings{wang2023enforcing,
	title = {Enforcing {Hard} {Constraints} with {Soft} {Barriers}: {Safe} {Reinforcement} {Learning} in {Unknown} {Stochastic} {Environments}},
	author = {Wang, Yixuan and Zhan, Simon Sinong and Jiao, Ruochen and Wang, Zhilu and Jin, Wanxin and Yang, Zhuoran and Wang, Zhaoran and Huang, Chao and Zhu, Qi},
    booktitle = {International Conference on Machine Learning},
	year = {2023}
}

@article{hsu2023safety,
  title={The safety filter: A unified view of safety-critical control in autonomous systems},
  author={Hsu, Kai-Chieh and Hu, Haimin and Fisac, Jaime F},
  journal={Annual Review of Control, Robotics, and Autonomous Systems},
  volume={7},
  year={2023},
  publisher={Annual Reviews}
}

@book{smyshlyaev2010adaptive,
  title={Adaptive control of parabolic PDEs},
  author={Smyshlyaev, Andrey and Krstic, Miroslav},
  year={2010},
  publisher={Princeton University Press}
}

@book{zhu2020delay,
  title={Delay-adaptive linear control},
  author={Zhu, Yang and Krstic, Miroslav},
  year={2020},
  publisher={Princeton University Press}
}

@book{karafyllis2025robust,
  title={Robust adaptive control: deadzone-adapted disturbance suppression},
  author={Karafyllis, Iasson and Krstic, Miroslav},
  year={2025},
  publisher={SIAM}
}

@article{degrave2022magnetic,
  title={Magnetic control of {T}okamak plasmas through deep reinforcement learning},
  author={Degrave, Jonas and Felici, Federico and Buchli, Jonas and Neunert, Michael and Tracey, Brendan and Carpanese, Francesco and Ewalds, Timo and Hafner, Roland and Abdolmaleki, Abbas and de Las Casas, Diego and others},
  journal={Nature},
  volume={602},
  number={7897},
  pages={414--419},
  year={2022},
  publisher={Nature Publishing Group UK London}
}

@article{bellemare2020autonomous,
  title={Autonomous navigation of stratospheric balloons using reinforcement learning},
  author={Bellemare, Marc G and Candido, Salvatore and Castro, Pablo Samuel and Gong, Jun and Machado, Marlos C and Moitra, Subhodeep and Ponda, Sameera S and Wang, Ziyu},
  journal={Nature},
  volume={588},
  number={7836},
  pages={77--82},
  year={2020},
  publisher={Nature Publishing Group UK London}
}

@article{da2025survey,
  title={A survey of sim-to-real methods in {RL}: {P}rogress, prospects and challenges with foundation models},
  author={Da, Longchao and Turnau, Justin and Kutralingam, Thirulogasankar Pranav and Velasquez, Alvaro and Shakarian, Paulo and Wei, Hua},
  journal={arXiv preprint arXiv:2502.13187},
  year={2025}
}

@article{hafner2025training,
  title={Training agents inside of scalable world models},
  author={Hafner, Danijar and Yan, Wilson and Lillicrap, Timothy},
  journal={arXiv preprint arXiv:2509.24527},
  year={2025}
}

@inproceedings{lawrence2025mpcritic,
  title={MPCritic: A plug-and-play {MPC} architecture for reinforcement learning},
  author={Lawrence, Nathan P and Banker, Thomas and Mesbah, Ali},
  booktitle={IEEE Conference on Decision and Control},
  pages={1048--1054},
  year={2025}
}

@article{hansen2023td,
  title={{TD-MPC2}: Scalable, robust world models for continuous control},
  author={Hansen, Nicklas and Su, Hao and Wang, Xiaolong},
  journal={arXiv preprint arXiv:2310.16828},
  year={2023}
}

@article{aminmansour2024mitigating,
  title={Mitigating Value Hallucination in {Dyna}-Style Planning via Multistep Predecessor Models},
  author={Aminmansour, Farzane and Jafferjee, Taher and Imani, Ehsan and Talvitie, Erin J and Bowling, Michael and White, Martha},
  journal={Journal of Artificial Intelligence Research},
  volume={80},
  pages={441--473},
  year={2024}
}

@article{van2019use,
  title={When to use parametric models in reinforcement learning?},
  author={Van Hasselt, Hado P and Hessel, Matteo and Aslanides, John},
  journal={Advances in Neural Information Processing Systems},
  volume={32},
  year={2019}
}

@article{pan2018organizing,
  title={Organizing experience: a deeper look at replay mechanisms for sample-based planning in continuous state domains},
  author={Pan, Yangchen and Zaheer, Muhammad and White, Adam and Patterson, Andrew and White, Martha},
  journal={International Joint Conference on Artificial Intelligence},
  year={2018}
}

@article{patterson2024empirical,
  title={Empirical design in reinforcement learning},
  author={Patterson, Andrew and Neumann, Samuel and White, Martha and White, Adam},
  journal={Journal of Machine Learning Research},
  volume={25},
  number={318},
  pages={1--63},
  year={2024}
}

@article{tsitsiklis1996analysis,
  title={Analysis of temporal-diffference learning with function approximation},
  author={Tsitsiklis, John and Van Roy, Benjamin},
  journal={Advances in Neural Information Processing Systems},
  volume={9},
  year={1996}
}

@inproceedings{thomas2014bias,
  title={Bias in natural actor-critic algorithms},
  author={Thomas, Philip},
  booktitle={International Conference on Machine Learning},
  pages={441--448},
  year={2014}
}

@article{graves2023off,
  title={Off-policy actor-critic with emphatic weightings},
  author={Graves, Eric and Imani, Ehsan and Kumaraswamy, Raksha and White, Martha},
  journal={Journal of Machine Learning Research},
  volume={24},
  number={146},
  pages={1--63},
  year={2023}
}

@article{patterson2022generalized,
  title={A generalized projected {B}ellman error for off-policy value estimation in reinforcement learning},
  author={Patterson, Andrew and White, Adam and White, Martha},
  journal={Journal of Machine Learning Research},
  volume={23},
  number={145},
  pages={1--61},
  year={2022}
}

@inproceedings{zhang2020provably,
  title={Provably convergent two-timescale off-policy actor-critic with function approximation},
  author={Zhang, Shangtong and Liu, Bo and Yao, Hengshuai and Whiteson, Shimon},
  booktitle={International Conference on Machine Learning},
  pages={11204--11213},
  year={2020}
}

@article{karmakar2018two,
  title={Two time-scale stochastic approximation with controlled Markov noise and off-policy temporal-difference learning},
  author={Karmakar, Prasenjit and Bhatnagar, Shalabh},
  journal={Mathematics of Operations Research},
  volume={43},
  number={1},
  pages={130--151},
  year={2018}
}

@article{gaur2024closing,
  title={Closing the gap: Achieving global convergence (last iterate) of actor-critic under {M}arkovian sampling with neural network parametrization},
  author={Gaur, Mudit and Bedi, Amrit Singh and Wang, Di and Aggarwal, Vaneet},
  journal={arXiv preprint arXiv:2405.01843},
  year={2024}
}

@inproceedings{chen2025convergence,
  title={On the convergence of continuous single-timescale actor-critic},
  author={Chen, Xuyang and Zhao, Lin},
  booktitle={International Conference on Machine Learning},
  year={2025}
}

@article{chan2022greedification,
  title={Greedification operators for policy optimization: Investigating forward and reverse {KL} divergences},
  author={Chan, Alan and Silva, Hugo and Lim, Sungsu and Kozuno, Tadashi and Mahmood, A Rupam and White, Martha},
  journal={Journal of Machine Learning Research},
  volume={23},
  number={253},
  pages={1--79},
  year={2022}
}

@inproceedings{haarnoja2018soft,
  title={Soft actor-critic: Off-policy maximum entropy deep reinforcement learning with a stochastic actor},
  author={Haarnoja, Tuomas and Zhou, Aurick and Abbeel, Pieter and Levine, Sergey},
  booktitle={International Conference on Machine Learning},
  pages={1861--1870},
  year={2018}
}

@article{kingma2014adam,
  title={Adam: A method for stochastic optimization},
  author={Kingma, Diederik P and Ba, Jimmy},
  journal={arXiv preprint arXiv:1412.6980},
  year={2014}
}

@inproceedings{silver2014deterministic,
  title={Deterministic policy gradient algorithms},
  author={Silver, David and Lever, Guy and Heess, Nicolas and Degris, Thomas and Wierstra, Daan and Riedmiller, Martin},
  booktitle={International Conference on Machine Learning},
  pages={387--395},
  year={2014},
}

@inproceedings{fujimoto2018addressing,
  title={Addressing function approximation error in actor-critic methods},
  author={Fujimoto, Scott and Hoof, Herke and Meger, David},
  booktitle={International Conference on Machine Learning},
  pages={1587--1596},
  year={2018}
}

@article{sutton1999policy,
  title={Policy gradient methods for reinforcement learning with function approximation},
  author={Sutton, Richard S and McAllester, David and Singh, Satinder and Mansour, Yishay},
  journal={Advances in Neural Information Processing Systems},
  volume={12},
  year={1999}
}

@article{dorfler2023data,
  title={Data-driven control: Part two of two: Hot take: Why not go with models?},
  author={D{\"o}rfler, Florian},
  journal={IEEE Control Systems Magazine},
  volume={43},
  number={6},
  pages={27--31},
  year={2023},
  publisher={IEEE}
}

@article{Auer2002UCB,
  title={{Finite-time Analysis of the Multi-armed Bandit Problem}},
  author={Auer, Peter and Cesa-Bianchi, Nicolo and Fischer, Paul},
  journal={Machine Learning},
  volume={47},
  pages={235--256},
  year={2002},
  publisher={Springer}
}

@inproceedings{kocsis2006uct,
  title={Bandit based {M}onte-{C}arlo planning},
  author={Kocsis, Levente and Szepesv{\'a}ri, Csaba},
  booktitle={European Conference on Machine Learning},
  year={2006},
  pages={282–-293}
}

@article{silver2016alphago,
  title={Mastering the game of {Go} with deep neural networks and tree search},
  author={Silver, David et al.},
  journal={Nature},
  volume={529},
  pages={484–-489},
  year={2016}
}

@article{silver2017alphazero,
  title={Mastering Chess and {S}hogi by self-play with a general reinforcement learning algorithm},
  author={Silver, David et al.},
  journal={arXiv preprint arXiv:1712.01815},
  year={2017}
}

@book{Khalil2002,
  title={{Nonlinear Systems}},
  author={Khalil, Hassan K.},
  year={2002},
  publisher={Prentice Hall},
  edition={3rd}
}

@book{goodwinMRAC1984,
title={Adaptive Filtering Prediction and Control},
author={Goodwin, G.C. and Sin, K.S.},
year={1984},
publisher={Prentice-Hall}
}

@book{sastry1989adaptive,
  title={Adaptive control: stability, convergence and robustness},
  author={Sastry, Shankar and Bodson, Marc},
  year={1989},
  publisher={Prentice Hall}
}

@book{anderson1986stability,
  title={Stability of adaptive systems: Passivity and averaging analysis},
  author={Anderson, Brian DO and Bitmead, Robert R and Johnson Jr, C Richard and Kokotovic, Petar V and Kosut, Robert L and Mareels, Iven MY and Praly, Laurent and Riedle, Bradley D},
  year={1986},
  publisher={MIT Press}
}

@book{Boyd1994,
  title={{Linear Matrix Inequalities in System and Control Theory}},
  author={Boyd, Stephen and El Ghaoui, Laurent and Feron, Eric and Balakrishnan, Venkataramanan},
  year={1994},
  publisher={SIAM}
}

@book{AndersonMoore1971,
  title={{Linear Optimal Control}},
  author={Anderson, Brian D. O. and Moore, John B.},
  year={1971},
  publisher={Prentice Hall}
}

@book{Bertsekas2017,
  title={{Dynamic Programming and Optimal Control}},
  author={Bertsekas, Dimitri P.},
  year={2017},
  publisher={Athena Scientific},
  edition={4th}
}

@article{astromHistory,
author="{\AA}str{\"o}m, Karl",
title="History of Adaptive Control",
journal="Encyclopedia of Systems and Control",
year="2021",
pages="902--909"
}

@article{annaswamy2021historical,
  title={A historical perspective of adaptive control and learning},
  author={Annaswamy, Anuradha M and Fradkov, Alexander L},
  journal={Annual Reviews in Control},
  volume={52},
  pages={18--41},
  year={2021},
  publisher={Elsevier}
}

@misc{caldwell1950control,
  title={Control system with automatic response adjustment},
  author={Caldwell, William I},
  year={1950},
  publisher={Google Patents},
  note={US Patent 2,517,081}
}

@article{gabor1961universal,
  title={A universal non-linear filter, predictor and simulator which optimizes itself by a learning process},
  author={Gabor, Dennis and Wilby, WPL and Woodcock, R},
  journal={Proceedings of the IEE-Part B: Electronic and Communication Engineering},
  volume={108},
  number={40},
  pages={422--435},
  year={1961}
}

@book{aircraftAdaptive1961,
title={Adaptive Control Systems},
author={E. Mishkin and L. Braun},
publisher={McGraw-Hill Publishing},
year={1961}
}

@article{rosenblatt1958perceptron,
  title={The perceptron: a probabilistic model for information storage and organization in the brain},
  author={Rosenblatt, Frank},
  journal={Psychological Review},
  volume={65},
  number={6},
  pages={386},
  year={1958}
}

@article{tang2025deep,
  title={Deep reinforcement learning for robotics: A survey of real-world successes},
  author={Tang, Chen and Abbatematteo, Ben and Hu, Jiaheng and Chandra, Rohan and Mart{\'\i}n-Mart{\'\i}n, Roberto and Stone, Peter},
  journal={Annual Review of Control, Robotics, and Autonomous Systems},
  volume={8},
  number={1},
  pages={153--188},
  year={2025},
  publisher={Annual Reviews}
}

@article{liu2022stability,
  title={Stability and control of power grids},
  author={Liu, Tao and Song, Yue and Zhu, Lipeng and Hill, David J},
  journal={Annual Review of Control, Robotics, and Autonomous Systems},
  volume={5},
  number={1},
  pages={689--716},
  year={2022}
}

@article{ma2023reinforcement,
  title={Reinforcement learning with model-based feedforward inputs for robotic table tennis},
  author={Ma, Hao and B{\"u}chler, Dieter and Sch{\"o}lkopf, Bernhard and Muehlebach, Michael},
  journal={Autonomous Robots},
  volume={47},
  number={8},
  pages={1387--1403},
  year={2023}
}

@article{elmkaiel2025embodied,
  title={Embodied Intelligence for Sustainable Flight: A Soaring Robot with Active Morphological Control},
  author={Elmkaiel, Ghadeer and Schmitt, Syn and Muehlebach, Michael},
  journal={npj Robotics},
  year={2026}
}

@article{he2025decision,
  title={Decision-dependent stochastic optimization: The role of distribution dynamics},
  author={He, Zhiyu and Bolognani, Saverio and D{\"o}rfler, Florian and Muehlebach, Michael},
  journal={arXiv preprint arXiv:2503.07324},
  year={2025}
}

@article{piazza2019century,
  title={A century of robotic hands},
  author={Piazza, Cristina and Grioli, Giorgio and Catalano, Manuel G and Bicchi, Antonio},
  journal={Annual Review of Control, Robotics, and Autonomous Systems},
  volume={2},
  number={1},
  pages={1--32},
  year={2019},
  publisher={Annual Reviews}
}

@article{Doyle1978,
  title={{Guaranteed Margins for LQG Regulators}},
  author={Doyle, John C.},
  journal={IEEE Transactions on Automatic Control},
  volume={23},
  number={4},
  pages={756--757},
  year={1978}
}

@book{powell2007approximate,
  title={{Approximate Dynamic Programming: Solving the Curses of Dimensionality}},
  author={Powell, Warren B.},
  year={2007},
  publisher={John Wiley \& Sons}
}

@book{bertsekas1996neuro,
  title={{Neuro-Dynamic Programming}},
  author={Bertsekas, Dimitri P. and Tsitsiklis, John N.},
  year={1996},
  publisher={Athena Scientific}
}

@article{lewis2009reinforcement,
  title={{Reinforcement Learning and Adaptive Dynamic Programming for Feedback Control}},
  author={Lewis, Frank L. and Vrabie, Draguna},
  journal={IEEE Circuits and Systems Magazine},
  volume={9},
  number={3},
  pages={32--50},
  year={2009}
}

@book{bellman1957dynamic,
  title={{Dynamic Programming}},
  author={Bellman, Richard},
  year={1957},
  publisher={Princeton University Press}
}

@article{sutton2002reinforcement,
  title={{Reinforcement Learning is Direct Adaptive Optimal Control}},
  author={Sutton, Richard S. and Barto, Andrew G. and Williams, Ronald J.},
  journal={IEEE Control Systems Magazine},
  volume={12},
  number={2},
  pages={19--22},
  year={2002}
}

@article{annaswamy2023adaptive,
  title={{Adaptive Control and Intersections with Reinforcement Learning}},
  author={Annaswamy, Anuradha M.},
  journal={Annual Review of Control, Robotics, and Autonomous Systems},
  volume={6},
  pages={65--93},
  year={2023}
}

@misc{sutton2019bitter,
  title={{The Bitter Lesson}},
  author={Sutton, Richard S.},
  howpublished={Incomplete Ideas (Blog)},
  url={http://www.incompleteideas.net/IncIdeas/BitterLesson.html},
  year={2019}
}

@article{sutton1988learning,
  title={Learning to Predict by the Methods of Temporal Differences},
  author={Sutton, Richard S.},
  journal={Machine Learning},
  volume={3},
  number={1},
  pages={9--44},
  year={1988},
  publisher={Springer}
}

@inproceedings{eberhard2023pink,
  title={Pink noise is all you need: Colored noise exploration in deep reinforcement learning},
  author={Eberhard, Onno and Hollenstein, Jakob and Pinneri, Cristina and Martius, Georg},
  booktitle={International Conference on Learning Representations},
  year={2023}
}

@article{muehlebach2025sample,
  title={The sample complexity of online reinforcement learning: A multi-model perspective},
  author={Muehlebach, Michael and He, Zhiyu and Jordan, Michael I},
  journal={International Conference on Learning Representations},
  year={2026}
}

@book{astrom2013adaptive,
  title={Adaptive Control},
  author={{\AA}str{\"o}m, Karl J. and Wittenmark, Bj{\"o}rn},
  edition={2nd},
  year={2013},
  publisher={Dover Publications}
}

@book{narendra1989stable,
  title={Stable Adaptive Systems},
  author={Narendra, Kumpati S. and Annaswamy, Anuradha M.},
  year={1989},
  publisher={Prentice Hall}
}

@book{ioannou2012robust,
  title={Robust Adaptive Control},
  author={Ioannou, Petros A. and Sun, Jing},
  year={2012},
  publisher={Dover Publications}
}

@book{krstic1995nonlinear,
  title={Nonlinear and Adaptive Control Design},
  author={Krsti{\'c}, Miroslav and Kanellakopoulos, Ioannis and Kokotovi{\'c}, Petar V.},
  year={1995},
  publisher={Wiley}
}

@article{dulac2021challenges,
  title={Challenges of Real-World Reinforcement Learning: Definitions, Benchmarks and Analysis},
  author={Dulac-Arnold, Gabriel and Levine, Nir and Mankowitz, Daniel J. and Li, Jerry and Paduraru, Cosmin and others},
  journal={Machine Learning},
  volume={110},
  number={9},
  pages={2419--2468},
  year={2021}
}

@article{xie2025safe,
  title={A Survey of Safe Reinforcement Learning Methods in Robotics},
  author={Xie, Yuen},
  journal={ITM Web of Conferences},
  year={2025}
}

@article{hewing2020cautious,
  title={Cautious Model Predictive Control Using {G}aussian Process Regression},
  author={Hewing, Lukas and Kabzan, Juraj and Zeilinger, Melanie N.},
  journal={IEEE Transactions on Control Systems Technology},
  volume={28},
  number={6},
  pages={2736--2743},
  year={2020}
}

@article{dean2020sample,
  title={Sample complexity of the linear quadratic regulator},
  author={Dean, Sarah and Mania, Horia and Matni, Nikolai and Recht, Benjamin and Tu, Stephen},
  journal={Foundations of Computational Mathematics},
  volume={20},
  pages={633--679},
  year={2020}
}

@article{ames2017control,
  title={Control barrier function based quadratic programs for safety critical systems},
  author={Ames, Aaron D. and Xu, Xiaojing and Grizzle, Jessy W. and Tabuada, Paulo},
  journal={IEEE Transactions on Automatic Control},
  volume={62},
  number={8},
  pages={3861--3876},
  year={2017}
}

@inproceedings{williams1992simple,
  title={Simple Statistical Gradient-Following Algorithms for Connectionist Reinforcement Learning},
  author={Williams, Ronald J.},
  booktitle={Advances in Neural Information Processing Systems},
  volume={5},
  year={1992}
}

@article{schulman2017proximal,
  title={Proximal Policy Optimization Algorithms},
  author={Schulman, John and Wolski, Filip and Dhariwal, Prafulla and Radford, Alec and Klimov, Oleg},
  journal={arXiv preprint arXiv:1707.06347},
  year={2017}
}

@inproceedings{todorov2012mujoco,
  title={{MuJoCo:} A physics engine for model-based control},
  author={Todorov, Emanuel and Erez, Tom and Tassa, Yuval},
  booktitle={IEEE/RSJ International Conference on Intelligent Robots and Systems},
  pages={5026--5033},
  year={2012}
}

@article{mnih2015human,
  author  = {Mnih, Volodymyr and Kavukcuoglu, Koray and Silver, David and Rusu, Andrei A. and Veness, Joel and Bellemare, Marc G. and Graves, Alex and Riedmiller, Martin and Fidjeland, Andreas K. and Ostrovski, Georg and Petersen, Stig and Beattie, Charles and Sadik, Amir and Antonoglou, Ioannis and King, Helen and Kumaran, Dharshan and Wierstra, Daan and Legg, Shane and Hassabis, Demis},
  journal = {Nature},
  number  = {7540},
  pages   = {529--533},
  title   = {Human-level control through deep reinforcement learning},
  volume  = {518},
  year    = {2015},
}

@inproceedings{eberhard-2025-partially,
  title = {Partially Observable Reinforcement Learning with Memory Traces},
  author = {Eberhard, Onno and Muehlebach, Michael and Vernade, Claire},
  booktitle = {International Conference on Machine Learning},
  year = {2025},
  volume = {267},
  pages = {14934--14949}
}

@inproceedings{sutton1991dyna,
  title={Dyna, an Integrated Architecture for Learning, Planning, and Reacting},
  author={Sutton, Richard S.},
  booktitle={AAAI Conference on Artificial Intelligence},
  pages={216--224},
  year={1991}
}

@article{tassa2018deepmind,
  title={Deepmind control suite},
  author={Tassa, Yuval and Doron, Yotam and Muldal, Alistair and Erez, Tom and Li, Yazhe and Casas, Diego de Las and Budden, David and Abdolmaleki, Abbas and Merel, Josh and Lefrancq, Andrew and others},
  journal={arXiv preprint arXiv:1801.00690},
  year={2018}
}

@book{mohriFoundations,
    author={Mehryar Mohri and Afshin Rostamizadeh and Ameet Talwalkar},
    title={Foundations of Machine Learning},
    publisher={MIT Press},
    year={2018},
    edition={second}
}

@article{zakka2025mujocoplayground,
      title={{MuJoCo} Playground},
      author={Kevin Zakka and Baruch Tabanpour and Qiayuan Liao and Mustafa Haiderbhai and Samuel Holt and Jing Yuan Luo and Arthur Allshire and Erik Frey and Koushil Sreenath and Lueder A. Kahrs and Carmelo Sferrazza and Yuval Tassa and Pieter Abbeel},
      year={2025},
      journal={arXiv preprint arXiv:2502.08844}
}

@inproceedings{osband2017posterior,
  title={Why is posterior sampling better than optimism for reinforcement learning?},
  author={Osband, Ian and Van Roy, Benjamin},
  booktitle={International Conference on Machine Learning},
  pages={2701--2710},
  year={2017}
}

@article{osband2016deep,
  title={Deep exploration via bootstrapped {DQN}},
  author={Osband, Ian and Blundell, Charles and Pritzel, Alexander and Van Roy, Benjamin},
  journal={Advances in Neural Information Processing Systems},
  volume={29},
  year={2016}
}

@article{martius2013information,
  title={Information driven self-organization of complex robotic behaviors},
  author={Martius, Georg and Der, Ralf and Ay, Nihat},
  journal={PLOS One},
  volume={8},
  number={5},
  pages={e63400},
  year={2013}
}

@article{bellemare2016unifying,
  title={Unifying count-based exploration and intrinsic motivation},
  author={Bellemare, Marc and Srinivasan, Sriram and Ostrovski, Georg and Schaul, Tom and Saxton, David and Munos, Remi},
  journal={Advances in Neural Information Processing Systems},
  volume={29},
  year={2016}
}

@article{jaksch10ucrl,
  author  = {Thomas Jaksch and Ronald Ortner and Peter Auer},
  title   = {Near-optimal Regret Bounds for Reinforcement Learning},
  journal = {Journal of Machine Learning Research},
  year    = {2010},
  volume  = {11},
  number  = {51},
  pages   = {1563--1600}
}

@inproceedings{zhao-2026-why,
  title = {Why Linear Recurrent Memory Works in Partially Observable Reinforcement Learning},
  author = {Zhao, Yike and Eberhard, Onno and Khammassi, Malek and Sayed, Ali H. and Muehlebach, Michael},
  booktitle = {International Conference on Machine Learning},
  year = {2026},
  volume = {306}
}

@inproceedings{eberhard-2026-commit,
  title = {Commit to the Bit: Reactive Reinforcement Learning Done Right},
  author = {Eberhard, Onno and Vernade, Claire and Muehlebach, Michael},
  booktitle = {International Conference on Machine Learning},
  year = {2026},
  volume = {306}
}

@inproceedings{littman1994memoryless,
  title         = {Memoryless policies: Theoretical limitations and practical results},
  author        = {Littman, Michael L},
  booktitle     = {International Conference on Simulation of Adaptive Behavior: From Animals to Animats},
  volume        = {3},
  pages         = {238--245},
  year          = {1994}
}

@article{krishnamurthy2016pac,
  title         = {{PAC} reinforcement learning with rich observations},
  author        = {Krishnamurthy, Akshay and Agarwal, Alekh and Langford, John},
  journal       = {Advances in Neural Information Processing Systems},
  volume        = {29},
  year          = {2016}
}

@article{papadimitriou,
 author = {Christos H. Papadimitriou and John N. Tsitsiklis},
 journal = {Mathematics of Operations Research},
 number = {3},
 pages = {441--450},
 title = {The Complexity of {M}arkov Decision Processes},
 volume = {12},
 year = {1987}
}

@inproceedings{golowich2023planning,
  author    = {Golowich, Noah and Moitra, Ankur and Rohatgi, Dhruv},
  title     = {Planning and Learning in Partially Observable Systems via Filter Stability},
  year      = {2023},
  booktitle = {ACM Symposium on Theory of Computing},
  pages     = {349--362}
}

@inproceedings{liu2022when,
  title     = {When Is Partially Observable Reinforcement Learning Not Scary?},
  author    = {Liu, Qinghua and Chung, Alan and Szepesvari, Csaba and Jin, Chi},
  booktitle = {Conference on Learning Theory},
  pages     = {5175--5220},
  year      = {2022},
  volume    = {178}
  }

@article{LaSze25,
	author = {Tor Lattimore and Csaba Szepesv\'ari},
	title = {Sample-Based Planning and Learning with Function Approximation},
	journal = {Statistical Sciences},
	number = 4,
	volume = 40,
	pages = {517--545}, 
	year = {2025},
}

@article{KuBe82,
  author = {P. R. Kumar and A. Becker}, 
  year = 1982, 
  title = {A new family of optimal adaptive controllers for {M}arkov chains}, 
  journal = {IEEE Trans. on Automatic Control}, 
  volume = 27, 
  pages = {137--146}
}

@article{LaiRobbins85,
title = {Asymptotically efficient adaptive allocation rules},
journal = {Advances in Applied Mathematics},
volume = {6},
number = {1},
pages = {4-22},
year = {1985},
issn = {0196-8858},
doi = {https://doi.org/10.1016/0196-8858(85)90002-8},
url = {https://www.sciencedirect.com/science/article/pii/0196885885900028},
author = {T.L Lai and Herbert Robbins}
}

@inproceedings{Strens00,
  author ={Strens, Malcolm J. A.},
  title = {A {B}ayesian framework for reinforcement learning},
  booktitle = {ICML}, 
  pages = {943--950}, 
  year = 2000
}

\newpage
\appendix
\section{Notation and correspondence between RL and control}

\label{app:notation}

Throughout this tutorial we adopt a unified notation whenever possible. Since RL and control theory have historically developed different terminologies and conventions, Table~\ref{tab:notation_correspondence} summarizes the main symbols used throughout the text and their correspondence with common notation in each community.

\begin{table}[h]
\centering
\caption{Correspondence between the notation adopted in this tutorial and common notation in RL and control theory.}
\label{tab:notation_correspondence}
\small
\begin{tabular}{llll}
\toprule
\textbf{Concept} & \textbf{Tutorial} & \textbf{Common RL} & \textbf{Common Control} \\
\midrule
State & $s_t$ & $s_t$ & $x_t$ \\
Action / control input & $a_t$ & $a_t$ & $u_t$ \\
Policy / controller & $\pi$ & $\pi$ & $\kappa$, $u=\kappa(x)$ \\
Parameterized policy & $\pi_\theta$ & $\pi_\theta$ & $\kappa_\theta$ \\
Dynamics & $f(s,a)$ or $p(\cdot|s,a)$ & $P$, $p$ & $f$, $A,B$ \\
Disturbance & $\eta_t$ & often implicit & $d_t$, $w_t$, $\eta_t$ \\
State space & $\mathcal{S}$ & $\mathcal{S}$ & $\mathcal{X}$ \\
Action space & $\mathcal{A}$ & $\mathcal{A}$ & $\mathcal{U}$ \\
Cost & $c(s,a)$ & $-r(s,a)$ & $\ell(x,u)$ \\
Reward & $r(s,a)$ & $r(s,a)$ & rarely used \\
Objective (min.) & $J(\pi)$ & uncommon & $J(\pi)$ \\
Objective (max.) & $V(\pi)$ & $J(\pi)$ & uncommon \\
Discount factor & $\gamma$ & $\gamma$ & often omitted \\
Value function & $V^\pi$ & $V^\pi$ & cost-to-go \\
Optimal value function & $V^\star$ & $V^\star$ & optimal cost-to-go \\
Action-value function & $Q^\pi$ & $Q^\pi$ & state-action cost-to-go \\
Lyapunov function & rarely used & $U$ & $V$ \\
Advantage function & $A^\pi$ & $A^\pi$ & rarely used \\
State distribution & $d^\pi$ & $d^\pi$ & invariant measure \\
Actor & $\pi_\theta$ & $\pi_\theta$ & parameterized controller \\
Critic & $q_w$ & $Q_w$, $Q_\phi$ & approximate cost-to-go \\
Reference signal & $r_t$ & goal signal & $r_t$ \\
Tracking error & $e_t$ & uncommon & $e_t$ \\
Feedback gain & $K$ & uncommon & $K$ \\
\bottomrule
\end{tabular}
\end{table}

\vspace{1em}

\end{document}